\documentclass[letterpaper]{article} 
\usepackage{aaai2026}  
\usepackage{times}  
\usepackage{helvet}  
\usepackage{courier}  
\usepackage[hyphens]{url}  
\usepackage{graphicx} 
\usepackage{natbib}  
\usepackage{caption} 
\usepackage[noend]{algpseudocode}
\usepackage{algorithmicx,algorithm}
\usepackage{subcaption}
\usepackage{multirow}
\usepackage{multicol}
\usepackage{colortbl}
\definecolor{mygray}{gray}{.9}
\usepackage{color}
\usepackage{xcolor} 
\usepackage{booktabs} 
\usepackage{amssymb}
 
\usepackage{newfloat}
\usepackage{listings}
\DeclareCaptionStyle{ruled}{labelfont=normalfont,labelsep=colon,strut=off} 
\floatstyle{ruled}
\newfloat{listing}{tb}{lst}{}
\floatname{listing}{Listing}
\title{Modality-Aware Bias Mitigation and Invariance Learning\\ for Unsupervised Visible-Infrared Person Re-Identification}
\author{
    Menglin Wang\textsuperscript{\rm 1},
    Xiaojin Gong\textsuperscript{\rm 2}\thanks{Corresponding author.},
    Jiachen Li\textsuperscript{\rm 2},
    Genlin Ji\textsuperscript{\rm 1} 
}
\affiliations{
    \textsuperscript{\rm 1}School of Computer and Electronic Information, Nanjing Normal University, China\\
    \textsuperscript{\rm 2}College of Information Science and Electronic Engineering, Zhejiang University, China\\
    lynnwang6875@gmail.com, gongxj@zju.edu.cn
}

\begin{document}

\maketitle

\begin{abstract}
Unsupervised visible-infrared person re-identification (USVI-ReID) aims to match individuals across visible and infrared cameras without relying on any annotation. Given the significant gap across visible and infrared modality, estimating reliable cross-modality association becomes a major challenge in USVI-ReID. Existing methods usually adopt optimal transport to associate the intra-modality clusters, which is prone to propagating the local cluster errors, and also overlooks global instance-level relations. By mining and attending to the visible-infrared modality bias, this paper focuses on addressing cross-modality learning from two aspects: bias-mitigated global association and modality-invariant representation learning. Motivated by the camera-aware distance rectification in single-modality re-ID, we propose modality-aware Jaccard distance to mitigate the distance bias caused by modality discrepancy, so that more reliable cross-modality associations can be estimated through global clustering. To further improve cross-modality representation learning, a `split-and-contrast' strategy is designed to obtain modality-specific global prototypes. By explicitly aligning these prototypes under global association guidance, modality-invariant yet ID-discriminative representation learning can be achieved. While conceptually simple, our method obtains state-of-the-art performance on benchmark VI-ReID datasets and outperforms existing methods by a significant margin, validating its effectiveness. 
\end{abstract}

\begin{links}
\link{Code}{https://github.com/Terminator8758/BMIL}
\end{links}

\section{Introduction}

Visible-infrared person re-identification~\cite{sysu17} is the task of matching the same person from visible to infrared camera or vice versa. Due to the potential application in night-time surveillance and person retrieval, this task has received increasing research interest. However, it is costly and time-consuming to annotate the cross-modality image relations, therefore unsupervised visible-infrared re-identification (USVI-ReID) has gained its popularity in recently years~\cite{H2H, ADCA, PGMAL, CCLNet, RPLR, yang2023GUR, SDCL, PCLMP, MMM, teng2024ULC}.

\begin{figure}[ht]
\centering
\begin{subfigure}{0.38\textwidth}
\centering
\includegraphics[width=1.0\textwidth]{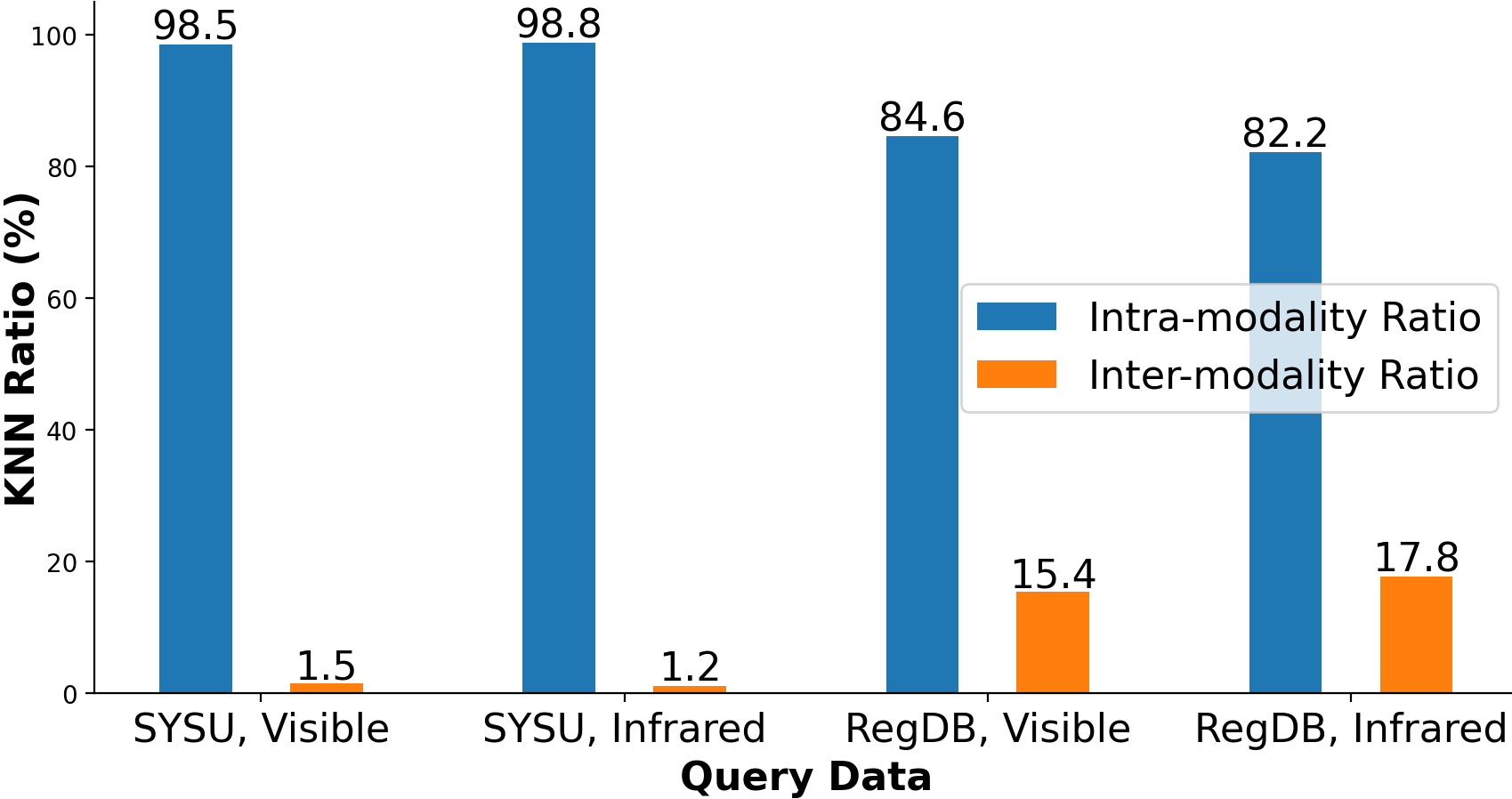} 
\caption{Intra and inter-modality ratio in Top-30 NNs}
\end{subfigure}
\\
\begin{subfigure}{0.34\textwidth}
\centering
\includegraphics[width=1.0\textwidth]{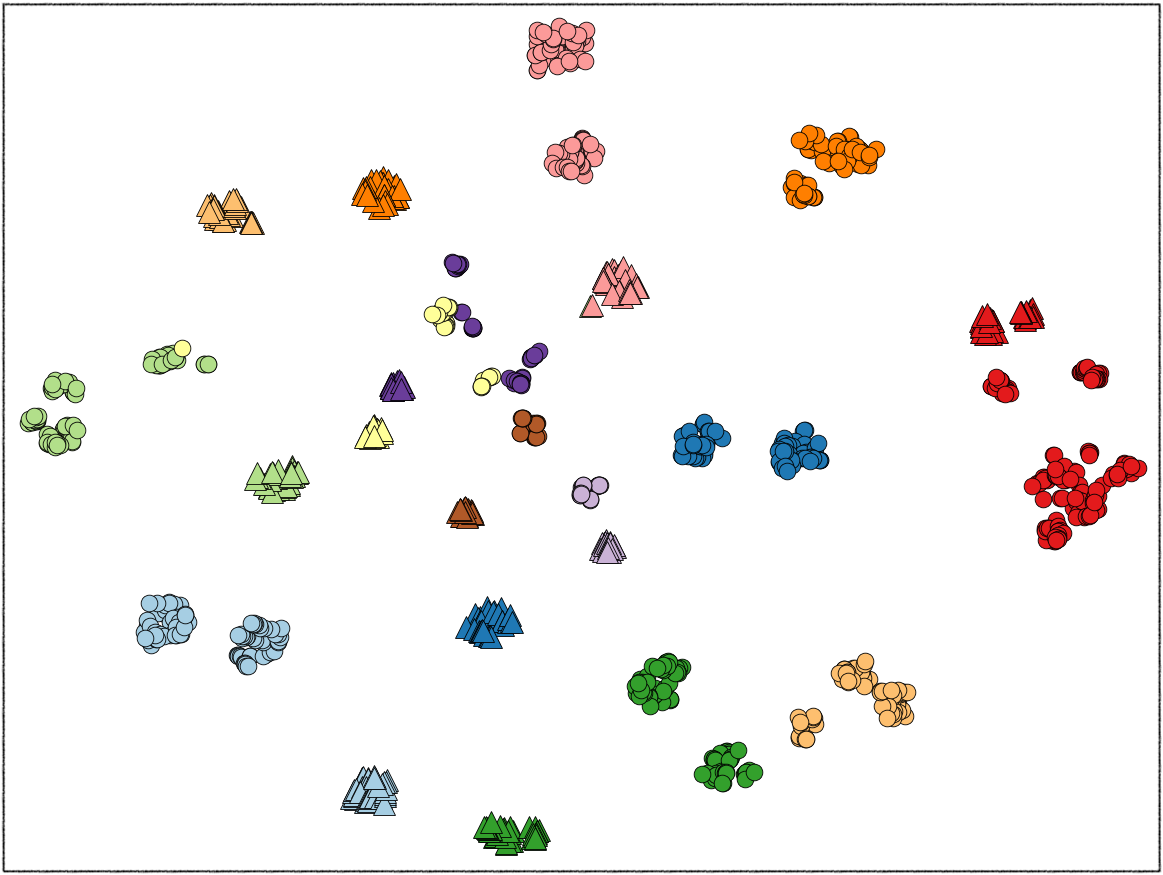} 
\caption{T-SNE visualization of visible (circle shape) and infrared (triangle shape) image features}
\end{subfigure}
\caption{Illustration of KNN distribution and feature visualization, using intra-modality (stage-1) trained model.}
\label{fig_intro}
\end{figure}

Given the substantial modality gap between visible and infrared images, most existing methods formulate USVI-ReID as a two-stage learning process. The first stage focuses on intra-modality learning to recognize identities from the same modality. This is typically done by iterative clustering and prototypical contrast learning, inspired by previous unsupervised single-modality re-ID methods~\cite{unsup_clustering, ge2020self, clustercontrast, Wang2021CAP, PPLR, MCRN, DCMIP}. The second stage deals with cross-modality matching and learning, where the key challenge is how to estimate reliable cross-modality association, under the presence of large modality discrepancy. A common strategy is to use optimal transport (e.g. Hungarian matching) to predict the correspondence among visible and infrared clusters, then enforce a cross-modality contrastive loss based on the matching result. However, this strategy propagates the intra-modality cluster errors while matching, and also overlooks the global instance relationship, reducing the efficiency of cross-modality learning.

On the other hand, a naive modality-agnostic global clustering is also not ideal. Due to the modality gap, intra-modality images generally exhibit higher similarity compared to cross-modality images, leading to a modality bias that negatively influences the pairwise distance distribution. Specifically, unsupervised clustering typically adopts pairwise Jaccard distance~\cite{Zhong2017reranking}, which entails the \textit{K}-nearest neighbor computation. With the modality bias, the retrieved KNNs are dominated by intra-modality instances, as observed in Figure \ref{fig_intro}(a). The biased neighbors further exacerbate the intra- and inter-modality distance gap, causing the global clustering to be unreliable and fail to associate cross-modality instances. Some recent efforts~\cite{10833701} try to select a larger number of KNNs and separately fuse visible and infrared features of the original KNNs. However, this does not address the inherently biased neighbor distribution, therefore has limited effect in rectifying the cross-modal distance.

To this end, we propose modality-aware global clustering to mitigate the modality bias and release the potential of global clustering. Inspired by the camera-aware distance rectification \cite{10657172} in single modality re-ID, we mitigate the bias in Jaccard distance by independently retrieving \textit{K}-nearest neighbors within each modality. These modality-balanced neighbors are incorporated into two key steps of Jaccard distance, \textit{i.e.} reciprocal neighbor computation and local query expansion, ensuring a fair contribution of both intra- and inter-modality neighbor instances in distance computation. As a result, the pairwise distance is modulated and less impacted by modality bias, thereby increasing the likelihood of associating cross-modality instances using global clustering.

While the cross-modality association bias is mitigated, there also exists issues in cross-modality representation learning. As identified by previous methods~\cite{DCMIP, PCLMP}, utilizing the centroid feature as the cluster prototype only represents the commonality, but fails to capture the variance and diversity of a cluster. The cluster variance becomes a more notable issue in cross-modality scenario, as the features of visible and infrared samples can have distinct distributions within the same global cluster. This is also supported by our visualizations in Figure \ref{fig_intro}(b), where images of the same modality are more likely to form a tight clique. Therefore, it is suboptimal to represent the global cluster with one unified centroid prototype. More informative prototypes are needed to precisely describe the modality-mixed cluster distribution.

Leveraging the modality label as a natural prior, we propose a \textit{split-and-contrast} strategy to split each cluster into modality-specific sub-clusters, and design modality-specific global prototypes to capture fine-grained dynamics within global clusters. By explicitly aligning the modality-specific prototypes using a multi-positive contrastive loss, the model is enforced to learn representations invariant to modality discrepancy. Together with the intra-modality prototypes, the representation learning is simultaneously guided by locally and globally associated prototypes, leading to a model that is robust to modality variation, yet discriminative at identity recognition.

To summarize, the contributions of this paper are:
\begin{itemize}
  \item{We present a simple yet effective method for USVI-ReID, targeting at cross-modality learning from the perspective of bias mitigation and invariance learning.}
  \item {To mitigate the distance bias caused by modality discrepancy, modality-aware Jaccard distance is proposed. Meanwhile, modality-specific global prototypes are designed to learn modality-invariant representations.}
  \item {On benchmark datasets, we perform extensive experiments and achieve state-of-the-art performance, demonstrating the effectiveness and superiority of our method.}
\end{itemize}

\section{Related Work}
\noindent \textbf{Supervised Visible-Infrared Re-ID.} The goal of supervised visible-infrared re-ID (VI-ReID) is to learn discriminative feature to recognize a person across modalities, by exploiting the given annotation. The VI-ReID task and the SYSU-MM01 dataset are initially proposed by Wu \textit{et. al}~\cite{sysu17}. Since cross-modality alignment is important for tackling the modality discrepancy, existing VI-ReID methods can be broadly categorized into image-level or feature-level alignment. For image-level alignment, D2RL~\cite{D2RL} generate cross-modality images using generative models. CAJ~\cite{CAJ} and CAJ+~\cite{caj_extend} design random channel exchange to generate color-irrelevant images, improving the robustness against color variations. For feature-level alignment, MACE~\cite{MACE} presents a middle-level sharable two-stream network to handle the modality discrepancy in both feature and classifier level. MAUM~\cite{9879066} utilizes historically accumulated proxies to align cross-modality features. PartMix~\cite{PartMix} randomly mixes cross-modality part-level features to regularize the model. FMCNet~\cite{FMCNet} and MUN~\cite{yu2023unify} generate intermediate modality feature to implicitly reduce the modality gap. Some recent methods~\cite{hu2024tvi, yu2025csdn} also exploit text modality to enhance visual feature learning. Although supervised VI-ReID performance has been significantly improved, the reliance on heavy annotation presents a limitation to its practical applicability.

\noindent \textbf{Unsupervised Visible-Infrared Re-ID.}
Due to the task characteristic, current USVI-ReID approaches usually combine intra-modality and cross-modality learning for method design. Driven by the success of single-modality unsupervised re-ID methods~\cite{clustercontrast, ge2020self, Wang2021CAP, PPLR, DCMIP}, intra-modality learning usually follows an iterative clustering and prototypical contrastive learning paradigm, to gradually boost the model performance. On this basis, some methods focus on intra-modality learning by designing multiple prototypes to represent cluster diversity or variance~\cite{PCLMP, MMM}, utilizing metric loss to assist representation learning~\cite{RPLR}, or considering neighbor information~\cite{teng2024ULC, RPLR} to refine the pseudo label. For cross-modality learning, the key issue is to predict reliable cross-modality association, and design efficient representation learning thereon. To this end, some methods estimate cross-modality correspondence by counting the number of reliable instance pairs in cross-modality clusters~\cite{ADCA}, designing label propagation based on normalized similarity~\cite{yang2023GUR}, exploiting optimal transport~\cite{OTLA} and graph matching~\cite{PGMAL}, or assigning instance weight according to cluster and neighbor correlation~\cite{teng2024ULC}. 

Despite the progress in USVI-ReID, one notable issue is the \textit{accumulation of cluster noise} when matching the intra-modality clusters. After associating the local clusters, intra-modality cluster noise is still kept and renders the association unreliable. What's more, associating the clusters ignores the global similarity relation among visible and infrared instances. In this paper, we aim to address this by proposing modality-aware global clustering to obtain more reliable association among instances.

\noindent \textbf{Debiased and Invariance Learning.}
Learning unbiased model from biased data has been a long-studied topic in visual recognition. Data bias such as image style, color or texture, can bring spurious correlations that negatively influence model learning. To address the bias, some methods~\cite{wang2021causal, wad2022equivariance} turn to causal intervention such as predicting data partition as effective backdoor adjustment. Regularization based debiasing methods are also popular for their lower complexity. For example, Invariant Risk Minimization (IRM)~\cite{arjovsky2019invariant} try to reduce the loss variance among different biases subsets, so as to achieve invariance to bias. EnD~\cite{tartaglione2021end} and BiasCon~\cite{hong2021unbiased} bring closer the representation of positive samples in case of different biases, while push apart negative samples sharing the same bias attribute. In general re-ID task, the camera bias also brings a spurious correlation to model. To reduce its influence in similarity measure, \cite{10657172} proposes camera-aware distance rectification that enhances the reliability of Jaccard distance. Compared to general re-ID, the visible-infrared modality discrepancy becomes a new source of bias in VI re-ID. In a similar spirit to~\cite{10657172}, we propose modality-aware Jaccard distance to effectively mitigate the bias and facilitate cross-modal clustering. Moreover, from the invariance learning perspective, we design a modality-invariant loss to explicitly optimize the modality-biased representations, thus better combat the modality confounder in model learning.

\section{Methodology}

\subsection{Overview}
\begin{figure*}[ht]
\centering
\includegraphics[width=0.85\textwidth]{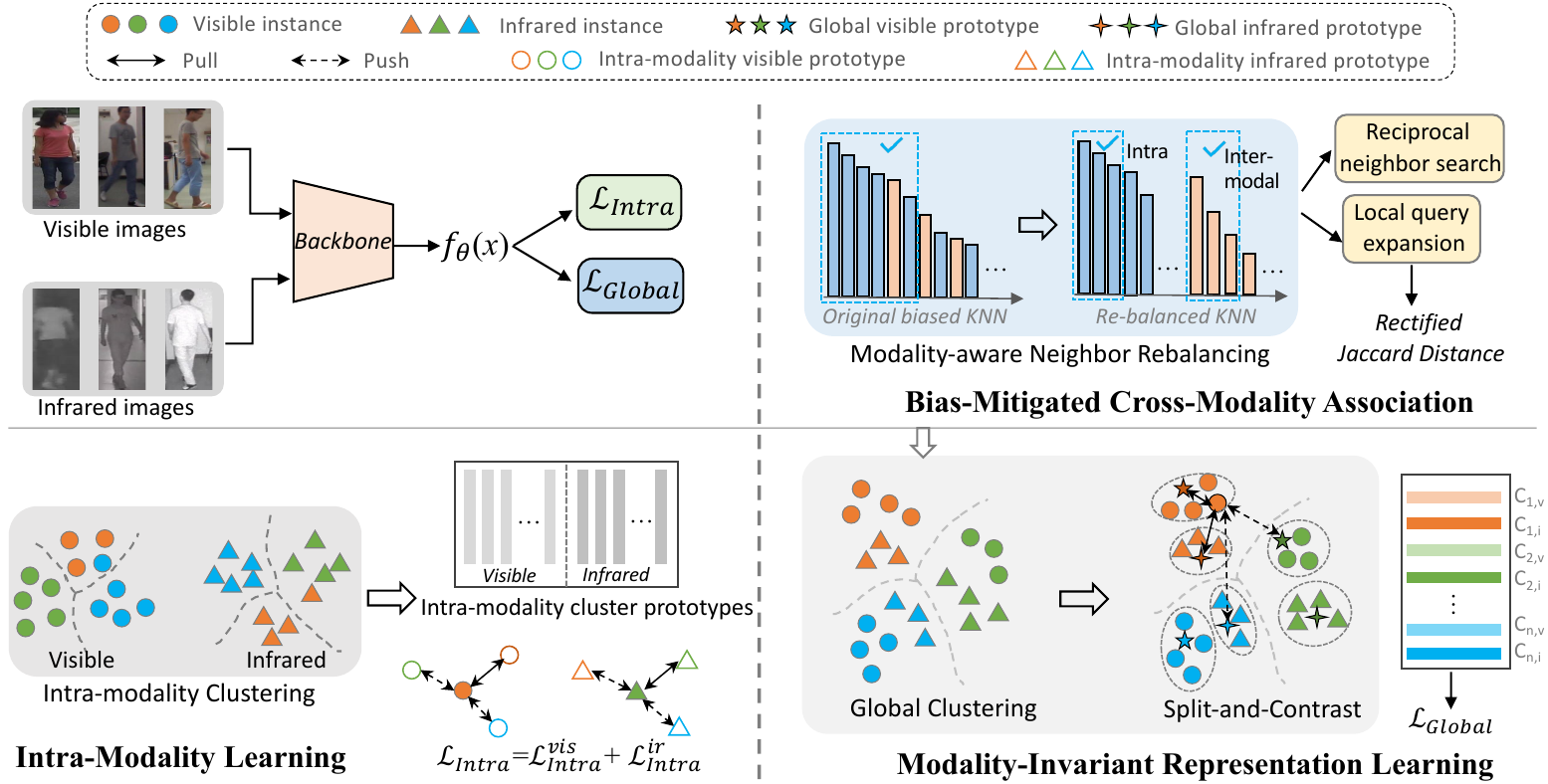}
\caption{An overview of the proposed modality-aware learning framework. For cross-modality learning, a bias-mitigating global association is proposed, featuring the modality-aware neighbor rebalancing strategy to rectify the pairwise distance and overcome the modality-induced bias. The rectified distance is utilized by global clustering to obtain modality-mixed clusters. A split-and-contrast design captures the variance of global clusters, then a multi-positive contrastive loss is designed to facilitate the learning of modality-invariant yet ID-discriminative representation.}
\label{fig_framework}
\end{figure*}

In USVI-ReID task, the overall aim is to learn a robust and discriminative model in an unsupervised setting, to match an identity across visible and infrared images. To achieve the goal, this paper addresses cross-modality learning from the perspective of mitigating the modality bias. Our method focuses on two complementary aspects: bias-mitigated global cross-modality association and modality-invariant representation learning. An overview of the proposed framework is illustrated in Figure \ref{fig_framework}. 
 The rest of this section is organized as follows: First, we introduce the preliminary information of the USVI-reID task, then we introduce the intra-modality learning. Afterwards, we present the cross-modality learning, where the proposed global association and invariant representation learning will be detailed.

\subsection{Intra-Modality Learning}
Suppose a cross-modality dataset is given as $\mathcal{D}=\{ \mathcal{D}_v, \mathcal{D}_r \}$, where $\mathcal{D}_v=\{x^v_1, x^v_2, ..., x^v_{N_1}\}$ denotes the visible training set composed of $N_1$ images, and $\mathcal{D}_r=\{x^r_1, x^r_2, ..., x^r_{N_2}\}$ is the infrared training set of $N_2$ images. As previous methods, we use a two-stream backbone $f_\theta$ to extract features for visible and infrared images. The backbone has separate initial convolution blocks for visible and infrared images, while the rest blocks are shared for both modalities.

Due to the image discrepancy in visible and infrared modalities, the backbone model is initially weak at matching cross-modality images. Therefore, a two-stage learning procedure is usually conducted, with intra-modality learning in the first stage to warm-up the model. For intra-modality learning, an iterative clustering and prototypical contrastive learning paradigm is adopted in each modality. 

Consider visible modality as an example. After obtaining a $d$-dimensional image feature $f_\theta(x^v)$ from the backbone $f_\theta$, DBSCAN clusters all visible images into $C^v$ clusters. Then for each cluster, a representative prototype is constructed and initialized as the cluster mean feature. A memory bank for visible modality is then formed as the collection of all clusters' prototypes, i.e. $M^v \in R^{C^v \times d}$. During mini-batch training, when the backbone parameter is updated by back-propagation, the prototype memory is updated by online image features:
\begin{equation}
	\mathcal{M}^v[j] \leftarrow \mu \mathcal{M}^v[j] + (1 - \mu) f_\theta(x^v_i),
	\label{eq:mu}
\end{equation}
where $\mathcal{M}^v[j]$ is the $j$-th entry of the memory bank, $\mu$ is the updating rate. $x^v_i$ is an image belonging to the $j$-th cluster.

To recognize different identities within the same modality, each visible training image $x^v_i$ is contrasted with all the visible prototypes using the InfoNCE loss~\cite{Gutmann2010NCE} as follow: 
\begin{equation}
\mathcal{L}^v_{intra} = - \sum_{i=1}^{N_b} \log \frac{exp(\mathcal{M}^v[\tilde{y}_i]^T f_\theta(x^v_i)/\tau)}{\sum_{j=1}^{C^v} exp(\mathcal{M}^v[j]^T f_\theta(x^v_i)/\tau)},
\label{eq_intra_loss}
\end{equation}
where $N_b$ is the batch size, $\tilde{y}_i$ is the pseudo label of image $x^v_i$ from clustering, $\tau$ is the temperature. When optimizing the prototypical contrastive loss in Eq. (\ref{eq_intra_loss}), the image is drawn closer to its belonging prototype, while pushed away from all the other prototypes, thus achieving discriminative representation learning in the same modality. 

For the infrared modality, the clustering, infrared prototype memory and the loss $\mathcal{L}^r_{intra}$ can be formulated similarly. The total intra-modality loss is then computed as: 
\begin{equation}
\mathcal{L}_{intra}=\mathcal{L}^v_{intra} + \mathcal{L}^r_{intra}.
\label{eq_intra_all_loss}
\end{equation}

\noindent \textbf{Dealing with over-clustering issue.} In real-world applications, it is common that a greater number of visible images are collected compared to infrared images. When applying merging-based clustering algorithms like DBSCAN on the visible modality, over-clustering tends to appear, \textit{i.e.} much more clusters are predicted than the actual identity number. To alleviate  over-clustering, we utilize a \textbf{\textit{subset clustering strategy}}~\cite{jin2022meta} to randomly sample a subset of images for clustering. By sampling at a fixed ratio (\textit{e.g.} 0.5), the average number of images per identity is reduced, making it easier to cluster the sampled images. In addition, each epoch samples a different subset of images due to randomness, making sure the full image set gets covered during training. Our experiments show that the subset clustering reduces over-clustering while also saves computation cost.

\subsection{Bias-Mitigated Cross-Modality Association}
Based on prototypes obtained from intra-modality clustering, a common way to establish cross-modality correspondence is to match the cross-modality prototypes through optimal transport algorithms such as Hungarian~\cite{Kuhn1955hungarian} or Sinkhorn algorithm~\cite{NIPS2013sinkhorn}. However, this two-step matching process is prone to accumulating the label noise of intra-modality clustering. Through cluster matching, intra-modality noise is falsely propagated across modality.

\subsubsection{The modality bias.} With the perception of instance-level similarity relation, one might consider global clustering as a suitable choice for cross-modality association. However, modality-induced bias becomes a main obstacle limiting the effectiveness of global clustering. The recognition of cross-modality images is confounded by the modality-specific context, such as image style, color and texture. As a result, a spurious correlation is formed between modality information and image similarity, negatively impacting the distance measure which is vital to clustering quality.  

To this end, we propose bias-mitigated cross-modality association by rectifying the pairwise distance of cross-modality images. For image clustering in typical unsupervised re-ID, Jaccard distance~\shortcite{Zhong2017reranking} is often adopted as the distance measure given its consideration of neighbor relations. In cross-modality scenario however, due to modality bias, the nearest neighbors are dominated by intra-modality images, as observed in Figure \ref{fig_intro}(a). The imbalanced neighbor composition only strengthens the interaction of intra-modality neighbors, but ignores the cross-modality ones, making Jacard distance weak in cross-modality association. Next, we revisit the key steps of Jaccard distance, then introduce our proposal for bias mitigation.

\subsubsection{Review of Jaccard distance.} 
Nearest neighbors play a crucial role in Jaccard distance computation. Given the features of all training images, $k_1$-nearest neighbors $N(x_i, k_1)$ are first retrieved for each query using Euclidean or Cosine distance. These KNNs are then involved in two key steps of Jaccard distance, \textit{i.e.} \textit{K}-reciprocal nearest neighbor retrieval (\textit{KRNN}) and local query expansion (\textit{LQE}).
 
(1) \textit{K-Reciprocal Nearest Neighbor Retrieval:} For sample $x_i$, its $k$-reciprocal nearest neighbor set is computed as: 
  $R(x_i,k_1)=\{x_j|x_j \in N(x_i, k_1) \wedge x_i \in N(x_j, k_1)\}$.  To increase the recall of positive samples, the reciprocal neighbors are expanded by adding the $\frac{1}{2}k_1$-reciprocal nearest neighbors of each candidate in $R(x_i,k_1)$. 
  
(2) \textit{Local Query Expansion:} To obtain a more robust feature, this step uses the $k_2$-nearest neighbors to compute an averaged feature as the local expansion result.

\subsubsection{Bias-mitigated distance calibration.} 
To rectify the biased distance and ensure fair contribution of intra and inter-modality neighbors, we propose a modality-aware distance rectification. Essentially, we consider a balanced number of modality-specific neighbors, and incorporate them into Jaccard distance. We do so by altering the following two basic steps of Jaccard distance computation:

(1) \textit{\textbf{Initial \textit{K}-nearest neighbor rectification.}} We independently search intra- and inter-modality $\frac{1}{2}k_1$-nearest neighbors. Denote $N^{intra}(x_i, k)$ and $N^{inter}(x_i, k)$ as the intra- and inter-modality $k$-nearest neighbors, respectively. The rectified $k_1$-nearest neighbor set is represented as: 
\begin{equation}
N^{\ast} (x_i, k_1)=N^{intra}(x_i, k_1/2) \cup N^{inter}(x_i, k_1/2), 
\end{equation}
and then sorted according to the query-to-neighbor distance: 
\begin{equation}
\overline{N}^{\ast}(x_i, k_1)=Sort(N^{\ast}(x_i, k_1)).
\end{equation}

The rectified neighbors $\overline{N}^{\ast}(x_i, k_1)$ are then adopted to retrieve $k$-reciprocal nearest neighbors.

(2) \textit{\textbf{Balanced local query expansion.}} Instead of averaging over the distance distribution of the $k_2$-nearest neighbors, we compute the expanded distance using the union of intra- and inter-modality $\frac{1}{2}k_2$-nearest neighbors:
\begin{equation}
\overline{Dist}(x_i)=\frac{1}{k_2}  { \sum_{j \in N^{\ast}(x_i, k_2) }} Dist(x_j).
\end{equation}

\subsubsection{Discussion.} A few recent methods~\cite{10833701,MMM} also consider an improved Jaccard distance for global clustering, but with limited gains. The key difference between their method and ours is that they only consider the modality imbalance in local query expansion. In contrast, our modality-balanced calibration strategy directly mitigates the biased KNN, achieving more thorough rectification of cross-modality distance. The distance distribution in Figure \ref{fig_dist} further validates the effect of our method. Besides, when camera label is available, our method can be complementary to CA-Jaccard~\cite{10657172}, allowing jointly handling modality and camera biases.

\begin{figure}[h]
\centering
\begin{subfigure}{0.221\textwidth}
\centering
\includegraphics[width=1.0\textwidth]{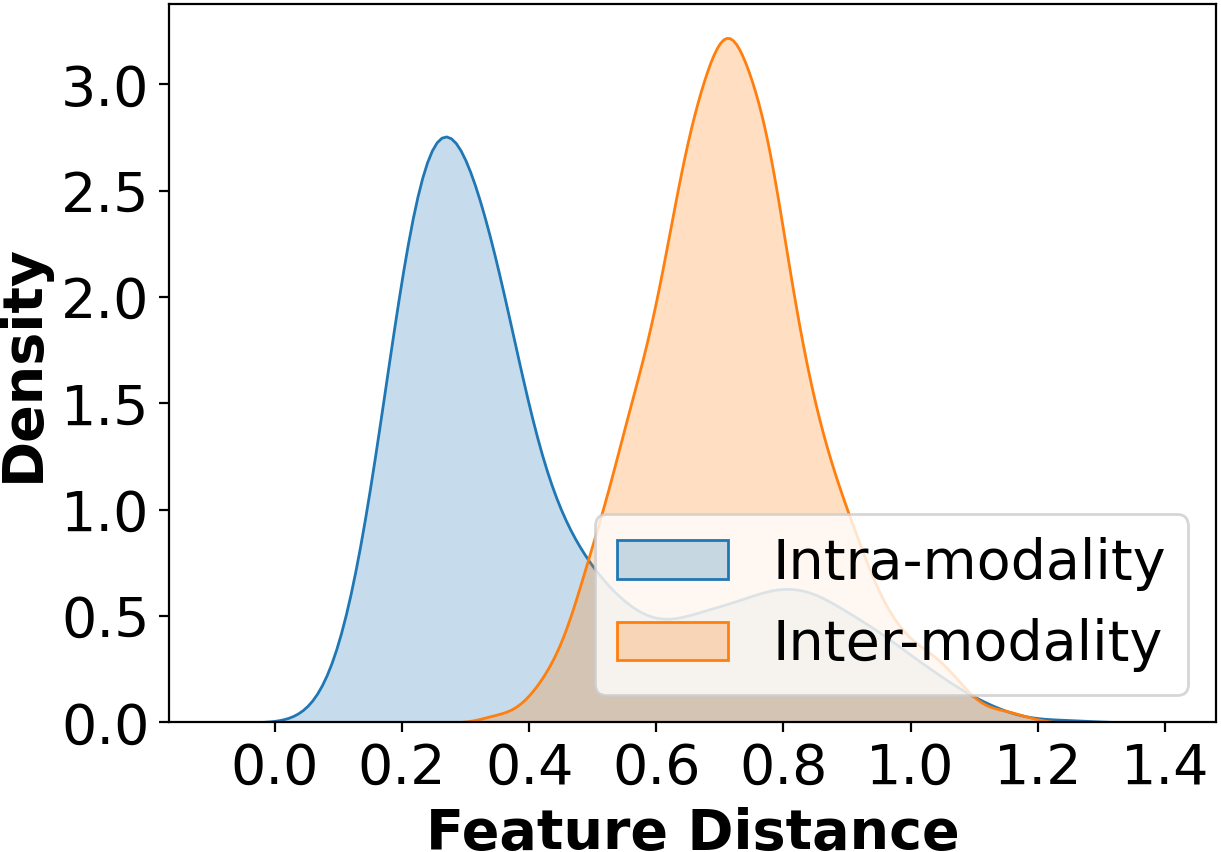} 
\caption{Cosine distance}
\end{subfigure}
\quad
\begin{subfigure}{0.221\textwidth}
\centering
\includegraphics[width=1.0\textwidth]{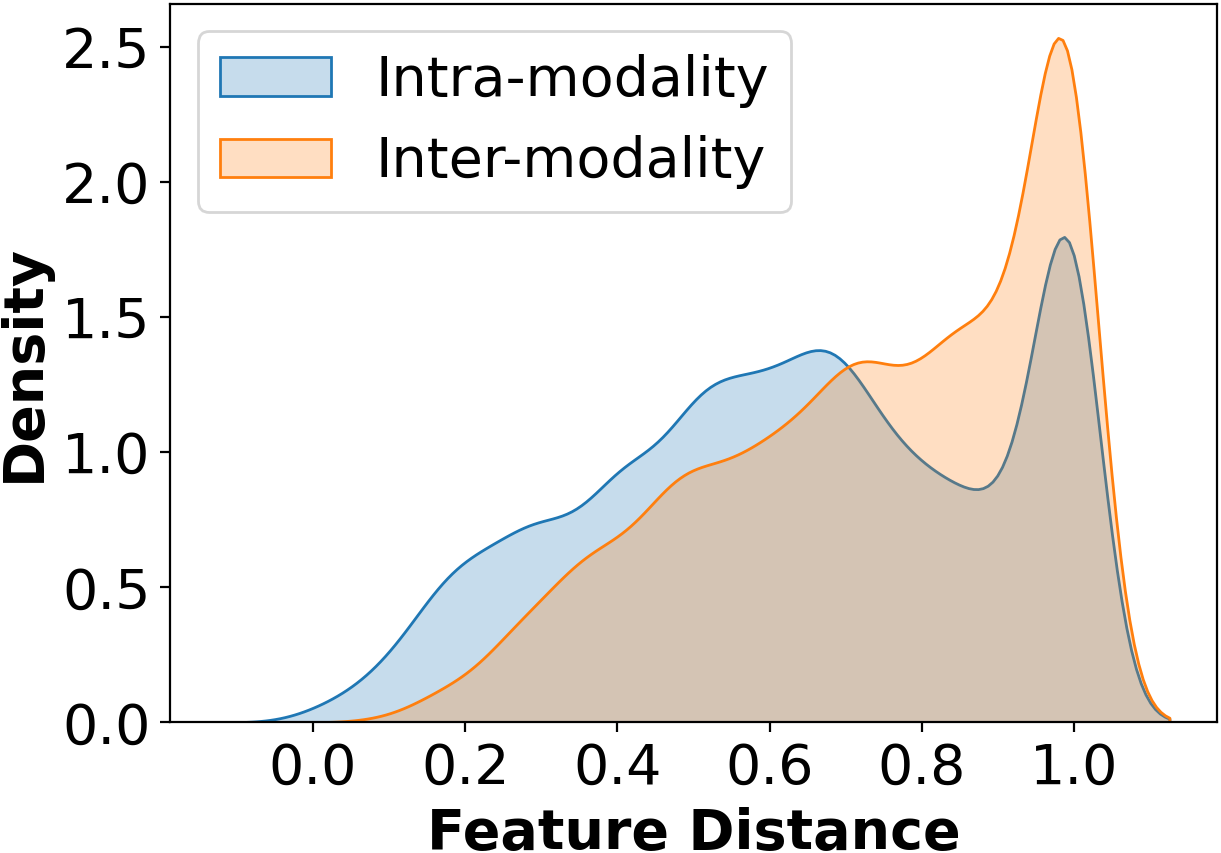} 
\caption{Our calibrated distance}
\end{subfigure}
\caption{ \small{Intra-cluster feature distance distribution on SYSU-MM01. Features are extracted with intra-modality trained model.}}
\label{fig_dist}
\end{figure}

\subsection{Modality-Invariant Representation Learning}
Unlike intra-modality clusters, the global clusters contain a mixture of visible and infrared images, therefore demonstrate a higher level of intra-cluster variance. As observed in Figure \ref{fig_intro}(b), images of different modalities form distinct cliques in feature space. Even when associated into the same cluster, the image features may still exhibit a ``\textit{two-peak}'' cosine distance distribution (Figure \ref{fig_dist}(a)).

\subsubsection{Modality-aware global prototypes.} 
 To preserve and model the intra-cluster characteristic, we leverage the available modality label as a natural prior to design modality-aware prototypes. Specifically, we propose a \textit{split-and-contrast} strategy to capture the fine-grained cluster distribution: Each global cluster is split into modality-specific subclusters, and a fine-grained prototype is associated to each subcluster. The prototypes are initialized as the subcluster mean feature, stored in a memory bank $\mathcal{K} \in R^{C\times d}$, and updated similarly as Eq. (\ref{eq:mu}). In this way, clusters containing mixed-modality images will be represented with two modality-specific global prototypes, preserving the variance of global clusters.

\subsubsection{Modality-invariant contrastive learning.} 
In a similar spirit to Invariant Risk Minimization~\cite{arjovsky2019invariant}, we would like to reduce the response variance of different modalities, so as to achieve invariance to modality bias. To this end, we propose a unified modality-invariant loss to jointly optimize the similarity relationship between query image and the modality-specific global prototypes. 

Our proposed loss generalizes InfoNCE to multi-positive setting. For a query visible image from a mixed global cluster, there exist two positive prototypes, one visible and one infrared. Using a multi-positive contrastive loss as below, we ensure the query feature to be equally close to its infrared and visible positive prototypes, thus explicitly aligning the modality-specific features for invariance learning:
\begin{equation}
\small
\mathcal{L}^v_{glb} = - \sum_{i=1}^{N_b} \frac{1}{|P(z_i)|} \sum_{p \in P(z_i)} \log \frac{exp(\mathcal{K}[p]^T f_\theta(x^v_i)/\tau)}{\sum\limits_{j \in S(x^v_i)} exp(\mathcal{K}[j]^T f_\theta(x^v_i)/\tau)},
\label{eq_fine_loss}
\end{equation}
where $z_i$ is the global cluster label of image $x^v_i$, and $P(z_i)$ is the index set of prototypes associated to cluster $z_i$. $S(x^v_i)=P(z_i) \cup NegK(x^v_i)$ is the index union of positive and Top-$k$ hard negative prototypes.

The global loss $\mathcal{L}^r_{global}$ for infrared modality can be formulated likewise. Then the average of the two losses is taken as the full global loss:
\begin{equation}
\mathcal{L}_{global} = 0.5 (\mathcal{L}^v_{glb} + \mathcal{L}^r_{glb}).
\label{eq_global_loss}
\end{equation}

\begin{table*}[ht]
\centering
\scalebox{0.83}{
\begin{tabular}{c|c|c||c|c|c|c||c|c|c|c}
\hline
\multicolumn{3}{c||}{\multirow{2}{*}{Settings}} & \multicolumn{4}{c||}{SYSU-MM01} & \multicolumn{4}{c}{RegDB} \\ \cline{4-11}
\multicolumn{3}{c||}{} & \multicolumn{2}{c|}{All Search} & \multicolumn{2}{c||}{Indoor Search} & \multicolumn{2}{c|}{Visible2Thermal} & \multicolumn{2}{c}{Thermal2Visible}\\
\hline
Type & Method & Venue & Rank-1 & mAP     & Rank-1 & mAP      & Rank-1 & mAP     & Rank-1 & mAP \\
\hline
\multirow{14}{*}{SVI-ReID}    
                ~ & DDAG~\cite{DDAG} & ECCV'20 & 54.8 & 53.0 & 61.0 & 68.0 & 69.4 & 63.5 & 68.1 & 61.8 \\ 
                ~ & AGW~\cite{AGW} & TPAMI'21 & 47.5 & 47.7 & 54.2 & 63.0 & 70.1 & 66.4 & 70.5 & 65.9 \\  
                ~ & CAJ~\cite{CAJ} & ICCV'21 & 69.9 & 66.9 & 76.3 & 80.4 & 85.0 & 79.1 & 84.8 & 77.8 \\  
                ~ & MPANet~\cite{MPANet} & CVPR'21 & 70.6 & 68.2 & 76.7 & 81.0 & 83.7 & 80.9 & 82.8 & 80.7 \\ 
                ~ & DART~\cite{DART} & CVPR'22 & 68.7 & 66.3 & 72.5 & 78.2 & 83.6 & 75.7 & 82.0 & 73.8 \\
                ~ & FMCNet~\cite{FMCNet} & CVPR'22 & 66.3 & 62.5 & 68.2 & 74.1 & 89.1 & 84.4 & 88.4 & 83.9 \\ 
                ~ & LUPI~\cite{LUPI} & ECCV'22 & 71.1 & 67.6 & 82.4 & 82.7 & 88.0 & 82.7 & 86.8 & 81.3 \\            
                ~ & DEEN~\cite{DEEN} & CVPR'23 & 74.7 & 71.8 & 80.3 & 83.3 & 91.1 & 85.1 & 89.5 & 83.4 \\
                ~ & PartMix~\cite{PartMix} & CVPR'23 & 77.8 & 74.6 & 81.5 & 84.4 & 85.7 & 82.3 & 84.9 & 82.5\\                 
                ~ & MUN~\cite{MUN} & ICCV'23 & 76.2 & 73.8 & 79.4 & 82.1 & 95.2 & 87.2 & 91.9 & 85.0\\     
                ~ & RLE~\cite{RLE} & NeurIPS'24 & 75.4 & 72.4  & 84.7 & 87.0 & 92.8 & 88.6 & 91.0 & 86.6\\   
                 ~ & CSDN~\cite{yu2025csdn} & TMM'25 & 75.2 & 71.8 & 82.0 & 85.0 & 89.0 & 84.7 & 88.2 & 82.8 \\
\hline
\multirow{3}{*}{SSVI-ReID}
               ~ & OTLA~\cite{OTLA} & ECCV'22 & 48.2 & 43.9 & 47.4 & 56.8 & 49.9 & 41.8 & 49.6 & 42.8\\
               ~ & TAA~\cite{taa} & TIP'23 & 48.8 & 42.3 & 50.1 & 56.0 & 62.2 & 56.0 & 63.8 & 56.5 \\
               ~ & DPIS~\cite{DPIS} & ICCV'23 & 58.4 & 55.6 & 63.0 & 70.0 & 62.3 & 53.2 & 61.5 & 52.7\\
\hline
\multirow{12}{*}{USVI-ReID}
               ~ & H2H~\cite{H2H} & TIP'21 & 30.2 & 29.4 & - & - & 23.8 & 18.9 & - & - \\ 
               ~ & OTLA~\cite{OTLA} & ECCV'22 & 29.9 & 27.1 & 29.8 & 38.8 & 32.9 & 29.7 & 32.1 & 28.6\\ 
               ~ & ADCA~\cite{ADCA} & MM'22 & 45.5 & 42.7 & 50.6 & 59.1 & 67.2 & 64.1 & 68.5 & 63.8\\
               ~ & NGLR~\cite{cheng2023unsupervised} & MM'23 & 50.4 & 47.4 & 53.5 & 61.7 & 85.6 & 76.7 & 82.9 & 75.0\\
               ~ & MBCCM~\cite{he2023efficient} & MM'23 & 53.1 & 48.2 & 55.2 & 62.0 & 83.8 & 77.9 & 82.8 & 76.7\\     
               ~ & CCLNet~\cite{CCLNet} & MM'23 & 54.0 & 50.2 & 56.7 & 65.1 & 69.9 & 65.5 & 70.2 & 66.7\\
               ~ & PGM~\cite{PGMAL} & CVPR'23 & 57.3 & 51.8 & 56.2 & 62.7 & 69.5 & 65.4 & 69.9 & 65.2\\           
               ~ & GUR~\shortcite{yang2023towards} & ICCV'23 & 61.0 & 57.0 & 64.2 & 69.5 & 73.9 & 70.2 & 75.0 & 69.9\\
               ~ & MMM~\cite{MMM} & ECCV'24 & 61.6 & 57.9 & 64.4 & 70.4 & 89.7 & 80.5 & 85.8 & 77.0\\
               ~ & PCLHD~\cite{PCLMP} & NeurIPS'24 & 64.4 & 58.7 & 69.5 & 74.4 & 84.3 & 80.7 & 82.7 & 78.4 \\
               ~ & PCLHD$\dagger$~\cite{PCLMP} & NeurIPS'24 & \underline{65.9} & \underline{61.8} & \underline{70.3} & \underline{74.9}  & 89.6 & 83.7 & 87.0 & 80.9\\
               ~ & RPNR~\cite{RPLR} & MM'24  & 65.2 & 60.0 & 68.9 &74.4 & \underline{90.9} & \underline{84.7}  & \underline{90.1} & \underline{83.2} \\
               \rowcolor{mygray}  
                ~ & Ours & - &  \textbf{67.1} & \textbf{63.1} & \textbf{75.0} & \textbf{78.6} & \textbf{94.3} & \textbf{89.1} & \textbf{93.6} & \textbf{88.5} \\
\hline
\end{tabular}
}
\caption{Comparison with state-of-the-art methods on SYSU-MM01 and RegDB dataset. PCLHD$\dagger$ denotes the combination of MMM and PCLHD. The best unsupervised performance is marked in \textbf{Bold} and the second best is marked in \underline{underline}.}
\label{compare_SOTA_table}
\end{table*}

\section{Experiments}

\subsection{Datasets and Settings}

\textbf{Datasets and evaluation protocols.} 
Following existing methods, we perform experiments on two benchmark VI-ReID datasets: SYSU-MM01~\cite{sysu17} and RegDB~\cite{regdb}. SYSU-MM01 is a large-scale dataset containing 491 identities captured by 6 cameras. The training set consists of 22,258 visible images and 11,909 infrared images from 395 identities. The rest 96 identities constitute the test set. RegDB contains 4,120 images from 412 identities, captured under one visible and one infrared camera. For RegDB, 10 random train/test splits are conducted, and the average accuracy of 10 splits is reported.

In line with the standard evaluation settings, Cumulative Matching Characteristics (CMC) and mean Average Precision (mAP) are utilized for accuracy measure. Specifically, Rank-1 accuracy is reported, along with mAP.

\noindent \textbf{Implementation details.}
Based on ResNet-50, AGW~\cite{AGW} is adopted as the backbone network and initialized with ImageNet-pretrained weights. Following existing methods, the training is conducted in two stages. The first stage focuses on intra-modality learning and the second stage further adds cross-modality learning. During training, Adam optimizer is utilized, with an initial learning rate of 3.5e-3 and divided by 10 every 20 epochs. Each stage is trained for 50 epochs. Batch size is 128 for both visible and infrared modality. Based on the clustering results per epoch, each mini-batch randomly samples 8 pseudo identities and 16 instances per identity. DBSCAN is utilized for clustering both within and cross-modality. Following PGM~\cite{PGMAL}, the \textit{eps} value for DBSCAN is set as 0.6 for SYSU-MM01 and 0.3 for RegDB. The memory update rate $\mu$ is set as 0.1, and temperature $\tau$ is 0.05. For more details and the pseudo Algorithm, please refer to the \textit{Appendix}.

\subsection{Comparison with the State-of-the-arts}
To validate the effectiveness of our proposed method, we compare with state-of-the-art unsupervised, semi-supervised and full-supervised VI-ReID methods on SYSU-MM01 and RegDB. The results are presented in Table \ref{compare_SOTA_table}. 

\noindent \textbf{Comparison with USVI-ReID methods.} As shown in Table \ref{compare_SOTA_table}, our method achieves superior accuracy compared to SoTA unsupervised methods while retaining its simplicity. On SYSU-MM01 dataset, our method improves the previous best Rank-1 Acc by $1.2\%$ and $4.7\%$ for \textit{All} and \textit{Indoor} search, respectively. On RegDB, our methods is also competitive, exceeding RPNR by $3.4\%$ and $4.8\%$ on average Rank-1 and mAP. The results demonstrate that the potential of global clustering has been overlooked by previous USVI-ReID methods. Bias-mitigated distance measure and invariant representation learning are both important aspects for enhancing the performance of USVI-ReID.

\noindent \textbf{Comparison with SVI-ReID and SSVI-ReID methods.} Supervised VI-ReID setting has been extensively studied over the last several years. From Table \ref{compare_SOTA_table}, we observe that USVI-ReID methods are catching up on the performance even without any annotation. Specifically, our proposed unsupervised method achieves comparable accuracy with some of the supervised VI-ReID methods, such as CAJ, DART and FMCNet. In addition, compared to the three semi-supervised VI-ReID methods listed in Table \ref{compare_SOTA_table}, our unsupervised method shows higher performance, proving the potential of learning from unlabeled data.

\subsection{Ablation Study}
In Table \ref{ablation_table}, we provide the ablation study to analyze the impact of each proposed component.

\begin{table}[ht]
	\centering
	\scalebox{0.88}{
	\begin{tabular}{c|ccc|cc|cc}
	\hline
                ~ &\multicolumn{3}{c|}{Components} & \multicolumn{2}{c|}{All Search} & \multicolumn{2}{c}{Indoor Search}\\
                \hline
               ~  & SC & BMGC & MIRL & R1 & mAP & R1 & mAP\\
               \hline
               $\mathcal{M}1$  & \multicolumn{3}{c|}{\textit{Intra Baseline}}                   & 39.5 & 38.9 & 47.1 & 55.5 \\
               $\mathcal{M}2$  & \multicolumn{3}{c|}{\textit{Global Baseline}}               & 54.9 & 51.5 & 62.9 & 68.9 \\     
               \hline    
               $\mathcal{M}3$  & ~ & \checkmark & ~                  & 64.9 & 60.0 & 68.0 & 73.5 \\
               $\mathcal{M}4$  & \checkmark & \checkmark & ~  & 64.8 & 61.0 & 73.9 & 77.4 \\
               $\mathcal{M}5$  & \checkmark  & ~  & \checkmark & 61.1 & 58.3 & 72.1 & 76.1 \\
               \rowcolor{mygray}  
               $\mathcal{M}6$  & \checkmark & \checkmark & \checkmark       & \textbf{67.1} & \textbf{63.1} & \textbf{75.0} & \textbf{78.6} \\
               \hline                
		\end{tabular}
        }
\caption{Ablation study on SYSU-MM01. `SC': Random Subset Clustering on visible modality. `BMGC': Bias-Mitigated Global Clustering. `MIRL': Modality-Invariant Representation Learning. `\textit{Intra Baseline'}: Intra-modality baseline. `\textit{Global Baseline}': Two-stage baseline with vanilla global clustering~\cite{PCLMP}.}
\label{ablation_table}
\end{table}

\noindent \textbf{Effectiveness of bias-mitigated global clustering.} Compared to the global baseline $\mathcal{M}2$, $\mathcal{M}3$ simply replaces the vanilla global clustering with our bias-mitigated global clustering, and gains a $10\%$ Rank-1 accuracy improvement in \textit{All} search mode. The significant accuracy boost validates the effectiveness of the improved global clustering. In addition, a comparison between $\mathcal{M}5$ and $\mathcal{M}6$ further proves the robustness of our global clustering when combined with other components, and highlights the importance of mitigating the modality bias for reliable cross-modality association.

\noindent \textbf{Effectiveness of modality-invariant representation learning.} By comparing $\mathcal{M}4$ with $\mathcal{M}6$, we observe that adding the modality-invariant representation learning brings further accuracy boost, improving the Rank-1 accuracy by $2.3\%$ in \textit{All} search. The results indicate that improving the cross-modality representation learning is indeed an aspect worth exploring. Our method captures the modality-induced cluster variance through a modality-invariant contrastive loss, so that better representation can be learned.

\noindent \textbf{Effectiveness of subset clustering.} By sampling on the visible modality, subset clustering mitigates the over-clustering as well as modality imbalance. From the comparison of $\mathcal{M}3$ and $\mathcal{M}4$, we see that utilizing the subset sampling strategy leads to clear improvement on the \textit{Indoor} search accuracy. This indicates that incorporating subset clustering is helpful for the recognition of basic and less challenging cases.

\noindent \textbf{Effectiveness of the full model.} Integrating all the proposed components, $\mathcal{M}6$ produces the best performance. Specifically, the accuracy boost compared to $\mathcal{M}1$ shows the overall effectiveness of our method for cross-modality learning. Compared to $\mathcal{M}3$-$\mathcal{M}5$, the full model's enhanced performance also proves the synergy of the proposed components.

\subsection{Parameter Analysis}
In this section, we delve into the performance of the proposed method by analyzing two aspects: clustering accuracy and feature visualization.

\begin{figure}[ht]
\centering
\includegraphics[width=0.405\textwidth]{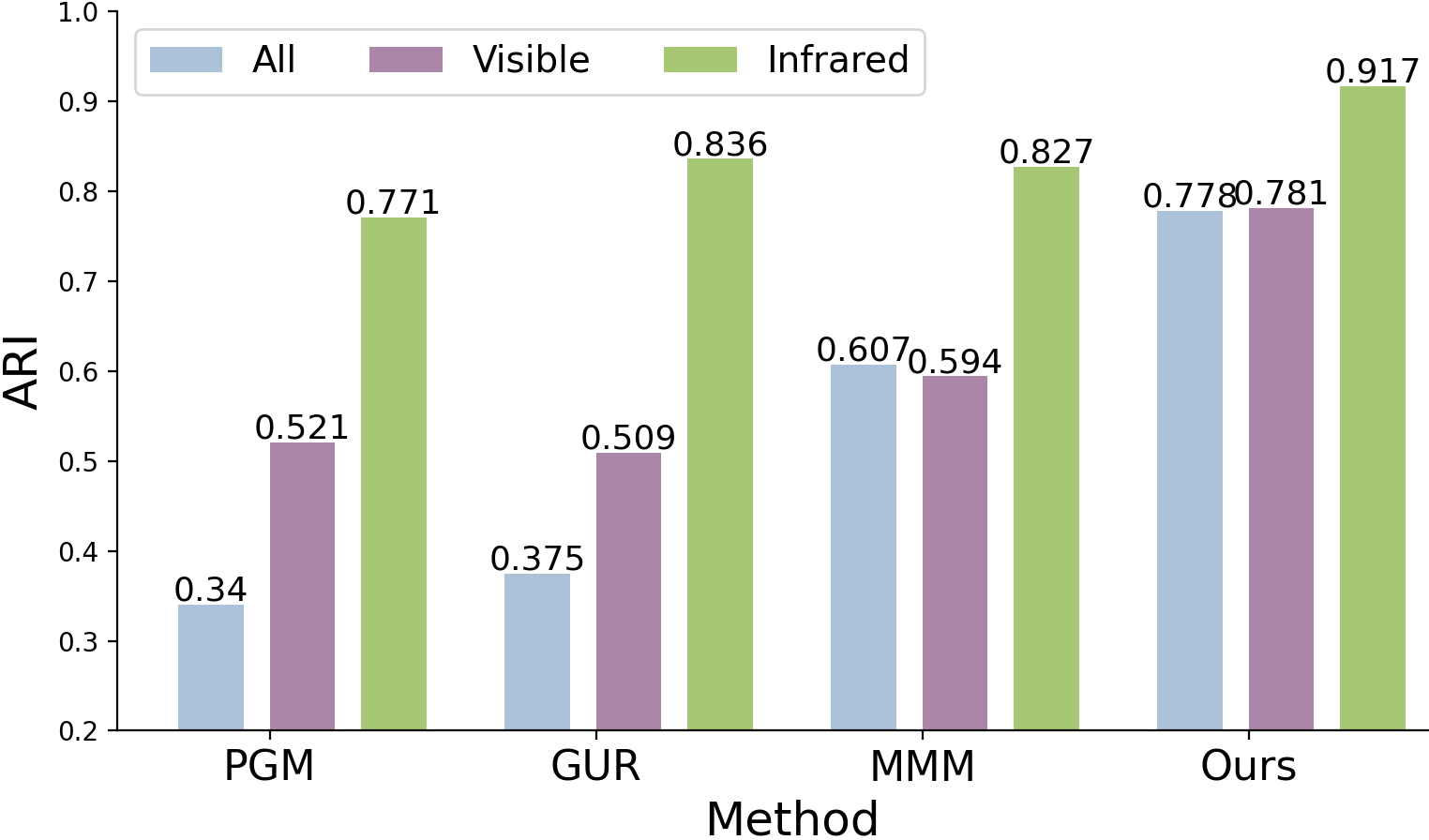}
\caption{Comparison of clustering accuracy ARI (Adjusted Random Index) on different methods.}
\label{fig_cluster_acc}
\end{figure}

\noindent \textbf{Analysis on clustering accuracy.}
The clustering accuracy is a direct reflection of association quality. To investigate how well our proposed global clustering performs, we present in Figure \ref{fig_cluster_acc} the cluster accuracy obtained by our method against other state-of-the-art methods. It can be observed that our method obtains a much higher ARI accuracy on both intra-modality and global clustering. The results prove our model is able to predict reliable pseudo labels for model learning.

\begin{figure}[ht]
\centering
\begin{subfigure}{0.2\textwidth}
\centering
\includegraphics[width=1.0\textwidth]{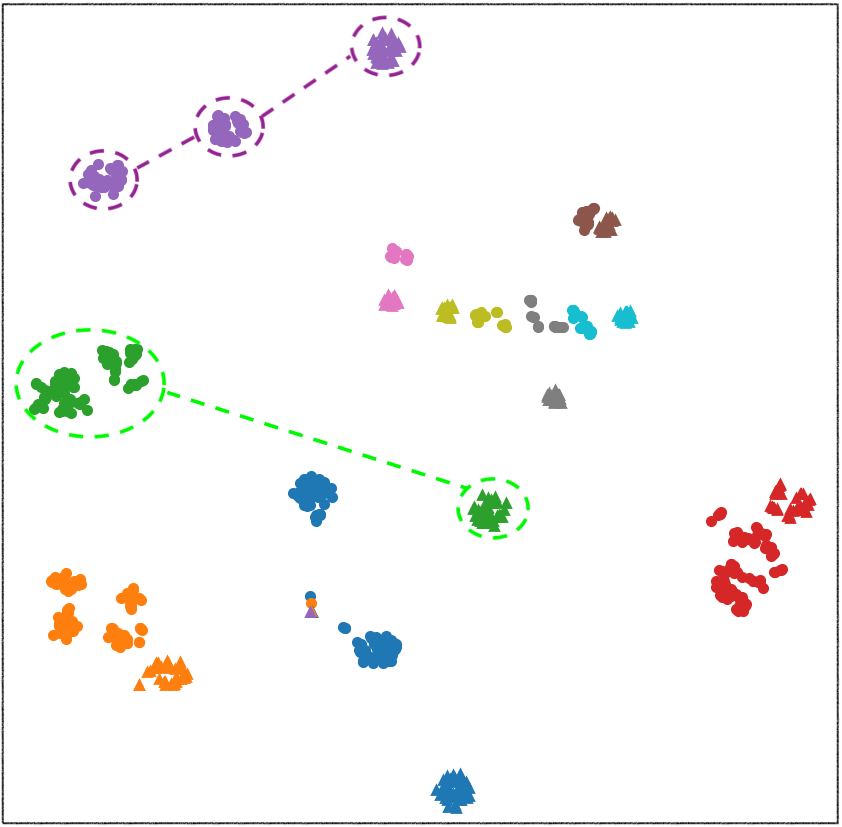} 
\caption{Baseline}
\end{subfigure}
\quad \quad
\begin{subfigure}{0.2\textwidth}
\centering
\includegraphics[width=1.0\textwidth]{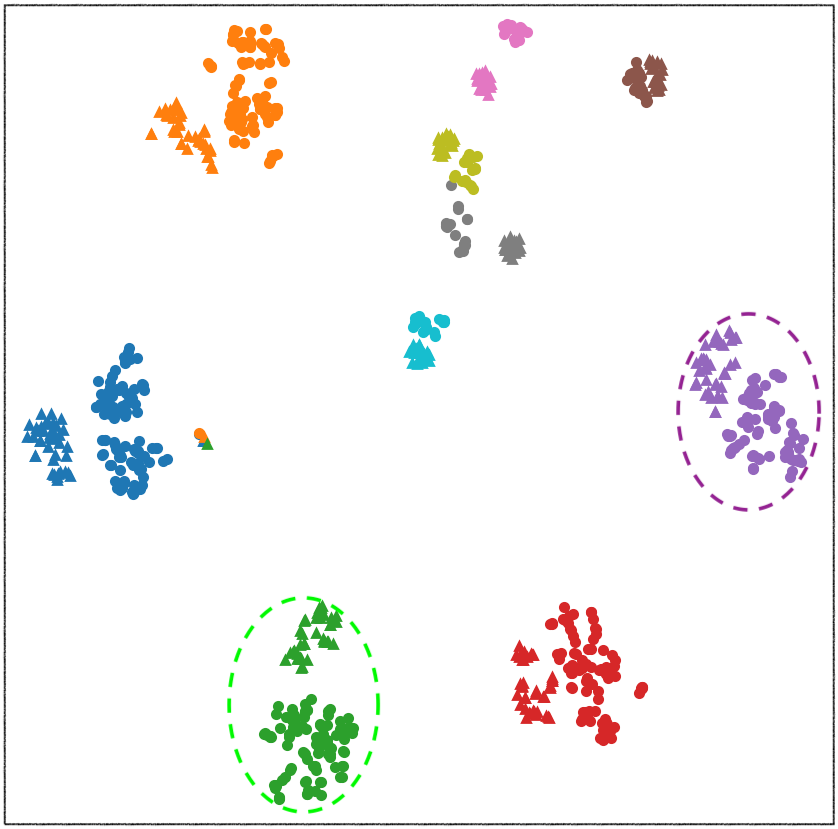} 
\caption{Our method}
\end{subfigure}
\caption{T-SNE visualization of features from 10 randomly selected identities in SYSU-MM01 training set. Different color represents different identity. Circle and triangle denote visible and infrared image, respectively.}
\label{fig_visualize}
\end{figure}

\noindent \textbf{Feature visualization. }
Figure \ref{fig_visualize} presents the feature visualization of our method compared to the global baseline. First, we observe that the baseline model produces multiple modality-specific clusters for the same identity, such as the purple and the green ones. This reflects that the baseline model is relatively weak at aggregating cross-modality images. As a comparison, our model generates compact clusters in most cases. With our model, samples of the purple identity are now gathered into the same cluster, so are the samples of the green identity. The visualization indicates that our model is able to enhance the intra-identity compactness while maintain the inter-identity separation.

\section{Conclusion}
In this paper, we propose a modality-aware bias mitigation and invariance learning method for USVI-ReID. To mitigate the bias in cross-modality distance measure, we propose modality-aware Jaccard distance so that reliable cross-modality association is obtained. To enhance the robustness to modality discrepancy, we design modality-specific global prototypes so that modality-invariant representation can be learned through a unified multi-positive contrastive loss. Despite our method's simplicity, it achieves the state-of-art performance on benchmark datasets, validating its effectiveness and showing the potential for further improvements in cross-modality learning.

\section*{Acknowledgements}
This work was supported by the Zhejiang Province Pioneer Research and Development Project “Research on Multi-modal Traffic Accident Holographic Restoration and Scene Database Construction Based on Vehicle-cloud Intersection” (No. 2024C01017), General Project of Basic Science Research in Higher Education Institutions of Jiangsu Province (No. 25KJD520008).

\bibliography{reference}


\setcounter{secnumdepth}{0} 

\section{A.1  More Implementation Details}
Following existing methods~\cite{PGMAL,MMM}, the network adopts a two-stream structure, where visible and infrared modality have separate initial Conv blocks, while the rest of backbone is shared. Data augmentation for training includes random flipping, cropping and erasing. On visible modality, random Grayscale augmentation is also utilized. 

When computing Jaccard distance, we adopt the same hyper-parameter setting for intra-modality and global clustering. That is, we set $k_1$ to 30 and $k_2$ to 6, as previous methods commonly do. Subset sampling is utilized for intra-modality clustering on SYSU-MM01, with sample ratio set to 0.5. Since RegDB is already a small dataset (with $2K$ images in each modality), there is no benefit in further sampling subset for clustering, so we use the original full set for clustering on RegDB. 
To jointly train with the intra-modality and global clustering results in the second training stage, a two-step update is utilized following~\cite{MMM, PCLMP}, computing the intra-modality loss and global loss on different batches. Specifically, an independent batch is sampled based on the global clustering result, and the global loss is computed on the global sampled batch. The detailed process is described in our Algorithm \ref{our_algo}. All experiments are conducted with Nvidia RTX 3090 using two GPUs.

\section{A.2  Pseudo Code of the Overall Algorithm} 
To enable better understanding of the proposed method, we provide the pseudo code of our method in Alg. \ref{our_algo}.

\begin{algorithm}[h]{
\caption{Our modality-aware learning algorithm}
\label{our_algo}
\hspace*{0.02in} {\bf Input:}
Unlabeled training set $\mathcal{D}$, network $f_{\theta}$\\
\hspace*{0.02in} {\bf Output:}
Trained network $f_{\theta}$
\begin{algorithmic}[1]
\For{epoch = 1 to \textit{maxEpoch}} \Comment{Stage 1}
    \State Perform intra-modality clustering;
    \State Construct prototype memory $\mathcal{M}^v$, $\mathcal{M}^r$;
    \State Generate the pseudo labeled dataset $\mathcal{D}'$;
    \For{batch = 1 to \textit{numBatch}}
        \State Sample a batch $\mathcal{B}' = \{(x_i^v, \tilde{y}_i^v)\cup (x_i^r, \tilde{y}_i^r)\}_{i=1}^{B}$;
        \State Compute the loss $\mathcal{L}_{intra}$ defined in Eq. (3);
        \State Backward to update network parameter $\theta$;
        \State Update prototype memory $\mathcal{M}^v$ and $\mathcal{M}^r$;
    \EndFor
\EndFor
\For{epoch = 1 to \textit{maxEpoch}}  \Comment{Stage 2}
    \State Perform intra-modality clustering;
    \State Compute rectified cross-modal Jaccard distance;
    \State Perform modality-aware global clustering;
    \State Construct intra-modal prototype memory $\mathcal{M}^v$, $\mathcal{M}^r$;
    \State Construct global prototype memory $\mathcal{K}$;
    \State Generate intra-modal pseudo labeled dataset $\mathcal{D}'$;
    \State Generate global pseudo labeled dataset $\mathcal{D}''$;
    \For{batch = 1 to \textit{numBatch}}
        \State Sample a batch $\mathcal{B}' = \{(x_i^v, \tilde{y}_i^v)\cup (x_i^r, \tilde{y}_i^r)\}_{i=1}^{B}$;
        \State Compute the loss $\mathcal{L}_{intra}$ defined in Eq. (3);
        \State Backward to update network parameter $\theta$;
        \State Update prototype memory $\mathcal{M}^v$ and $\mathcal{M}^r$;
         \State Sample a global mini-batch $\mathcal{B}'' = \{(x_i, z_i)\}_{i=1}^{B}$;
        \State Compute the loss $\mathcal{L}_{global}$ defined in Eq. (8);
        \State Backward to update network parameter $\theta$;
        \State Update global prototype memory $\mathcal{K}$ ;
    \EndFor
\EndFor
\end{algorithmic}
}
\end{algorithm}

\section{A.3  Computation Complexity Analysis}
Our method involves both intra-modality clustering and global clustering, where the computational bottleneck is the pairwise Jaccard distance. Suppose M and N are the number of visible and infrared images respectively. For intra-modality clustering, the time complexity is $O(M^2logM)$ for visible modality and $O(N^2logN)$ for infrared modality, similar to other methods. With subset clustering, the time complexity is reduced to $O(R^2M^2log(MR))$, where $\textit{R}$ is the sampling ratio (eg. 0.5).

\begin{table}[hb!]
\centering
\scalebox{0.85}{
\begin{tabular}{c|cc}
\toprule
Clustering & Jaccard distance time & Total time\\    
\midrule
Infrared modality& 21s &  22s  \\
Visible modality & 42s &  45s \\   
Visible modality(subset) & 20s &  21s \\   
Vanilla global & 47s &  51s \\   
Our global & 68s &  72s \\   
\bottomrule                
\end{tabular}
}
\caption{Time comparison of different clusterings.}
\label{table_time_compare}
\end{table}

For global clustering, the time complexity of our modality-aware Jaccard distance is $O((M+N)(MlogM+NlogN))$, compared to the vanilla Jaccard distance’s $O((M+N)^2log(M+N))$.

\begin{figure*}[ht!]
\centering
\begin{subfigure}{0.238\textwidth}
\centering
\includegraphics[width=1.0\textwidth]{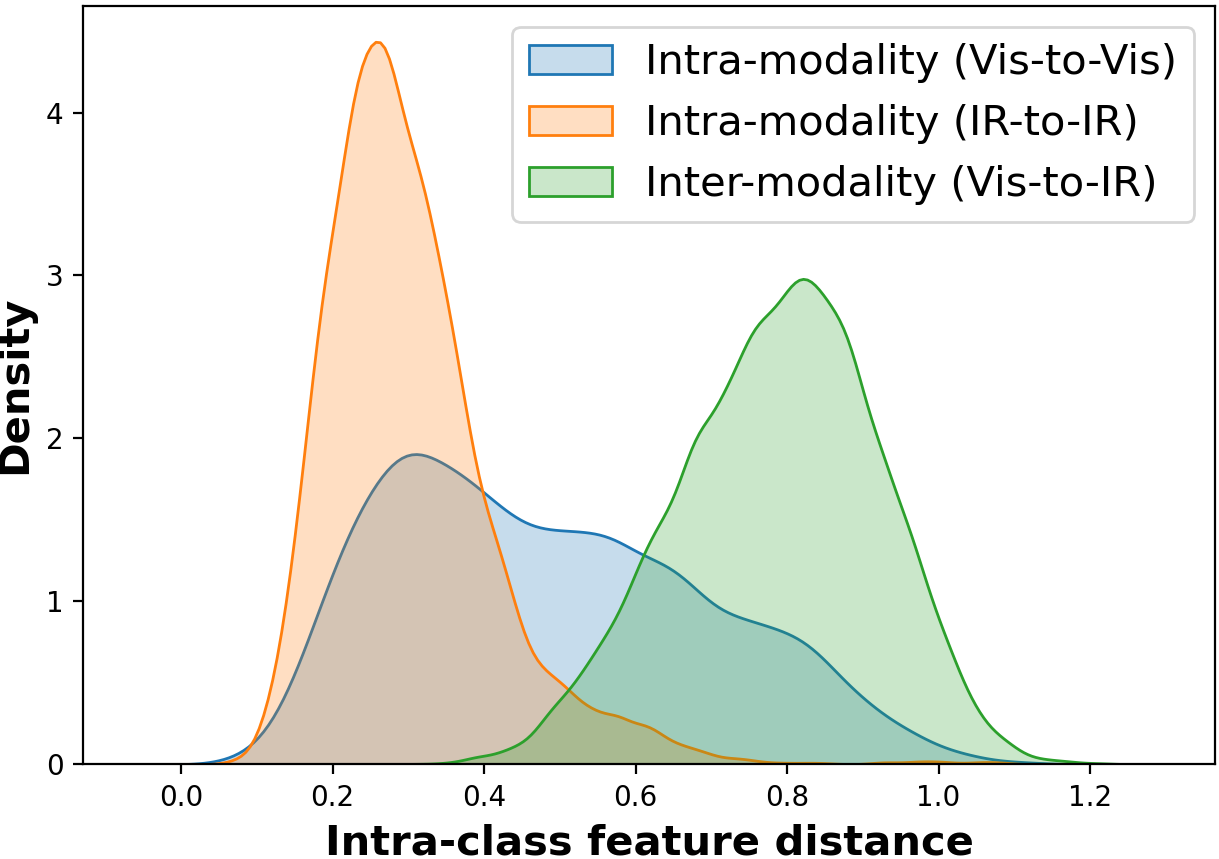} 
\caption{Cosine dist}
\end{subfigure}
\hspace{0.1cm}
\begin{subfigure}{0.237\textwidth}
\centering
\includegraphics[width=1.0\textwidth]{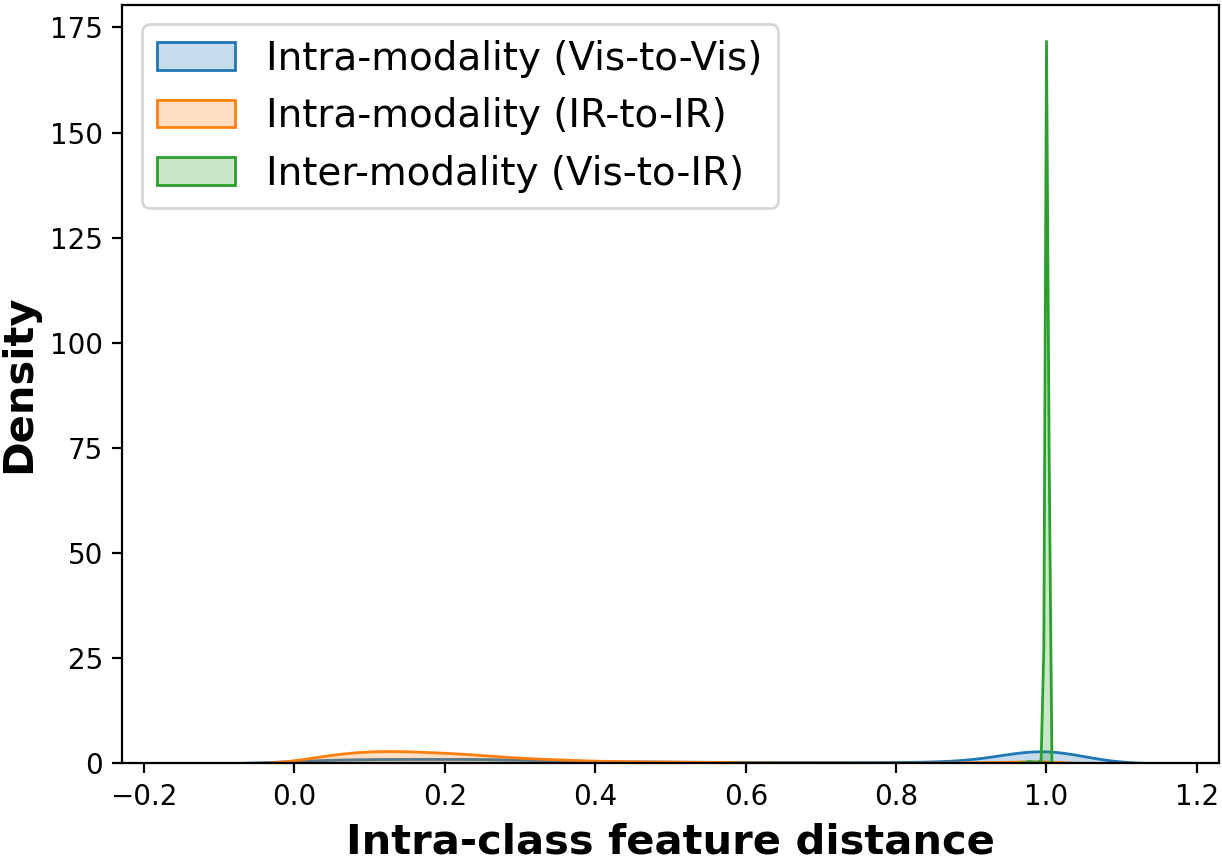} 
\caption{Vanilla Jaccard dist}
\end{subfigure}
\hspace{0.1cm}
\begin{subfigure}{0.238\textwidth}
\centering
\includegraphics[width=1.0\textwidth]{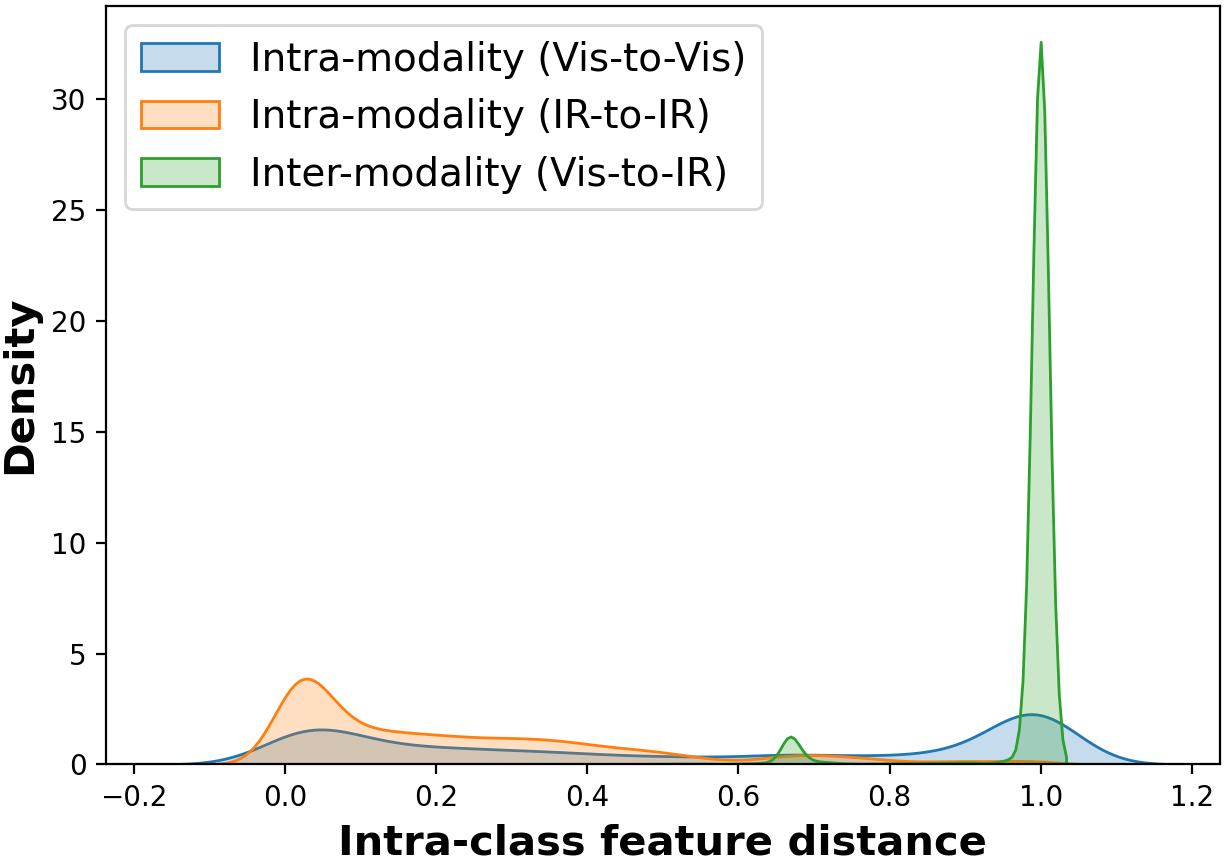} 
\caption{Improved Jaccard dist}
\end{subfigure}
\hspace{0.1cm} 
\begin{subfigure}{0.238\textwidth}
\centering
\includegraphics[width=1.0\textwidth]{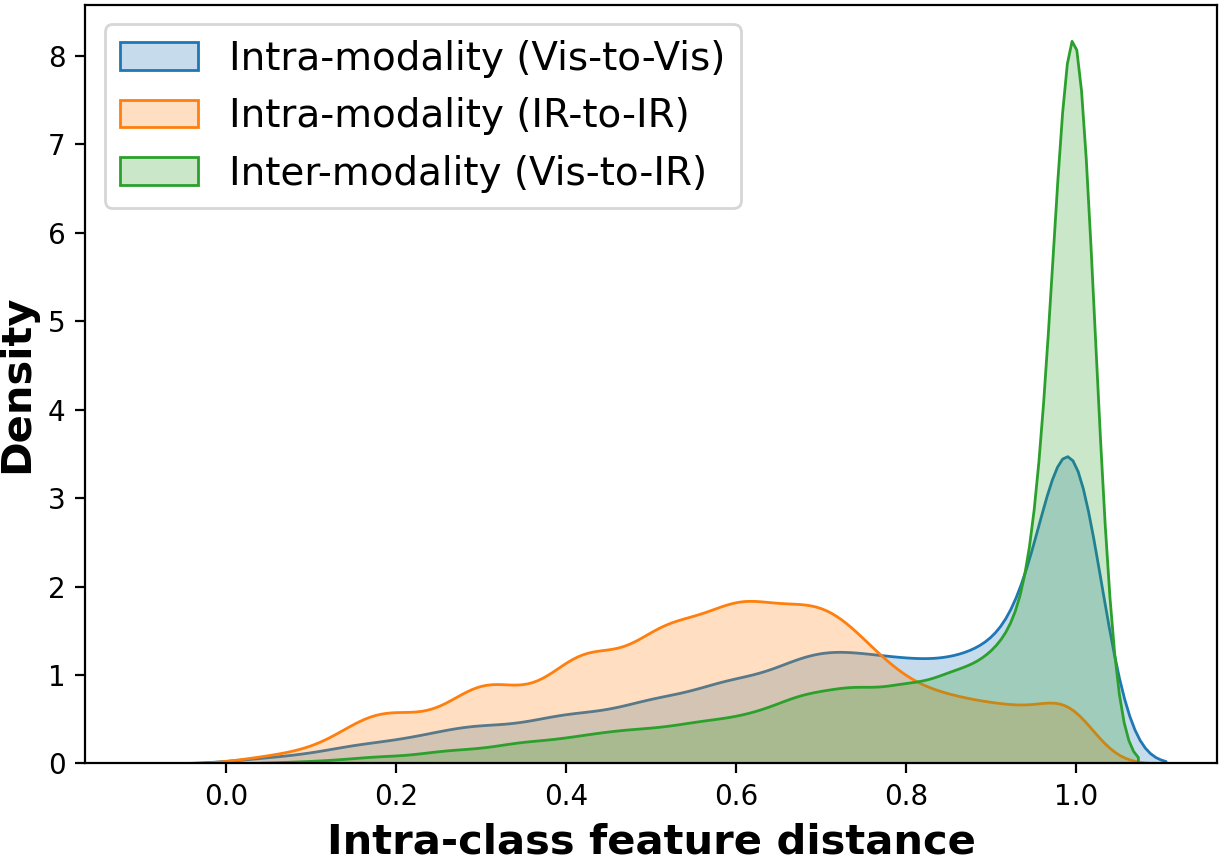} 
\caption{Our rectified dist}
\end{subfigure}
\caption{Intra-modality (Vis-to-Vis, IR-to-IR) and inter-modality (Vis-to-IR) distance distribution within the \textit{ground truth} classes. Improved Jaccard dist: the improved Jaccard distance proposed by~\cite{10833701}. Vis-to-Vis: visible-to-visible image distance. IR-to-IR: infrared-to-infrared. Vis-to-IR: visible-to-infrared. }
\label{fig_dist}
\end{figure*}

In our experiments on SYSU-MM01 using Nvidia-3090, the average computation time for different clusterings (running once) are summarized in Table \ref{table_time_compare}.

\section{A.4  More Experimental Results}

\subsection{A4.1  Analysis on Distance Rectification}
Recall that in Figure 3 of our main manuscript, the intra-cluster distance distribution is computed \textit{w.r.t.} the predicted clusters to demonstrate the intra-cluster variance caused by modality bias, and the mitigation effect of our method. 

To gain a comprehensive view of our proposed distance rectification, Figure \ref{fig_dist} here gives a more detailed comparison of the intra-class distance distribution using different metrics. Specifically, the distributions are computed within the \textit{ground truth} classes. By observing Figure \ref{fig_dist}(a), we see a clear margin between intra-modality and cross-modality mean distance, indicating that the modality bias is significant when using Cosine distance. Due to the biased cosine distance, the vanilla Jaccard distance also fails to improve the distance gap, and the inter-modality distance distribution becomes more concentrated. In Figure \ref{fig_dist}(c), the improved Jaccard distance~\cite{10833701} slightly mitigates the situation, but still the intra and inter-modality distribution gap is large and harms cross-modality association. In contrast, our proposed distance rectification produces much improved distance distribution, largely reducing the distance gap, making it easier to associate cross-modality samples.

\subsection{A4.2  Visual Results from Clustering}

\begin{figure}[h]
\centering
\includegraphics[width=0.47\textwidth]{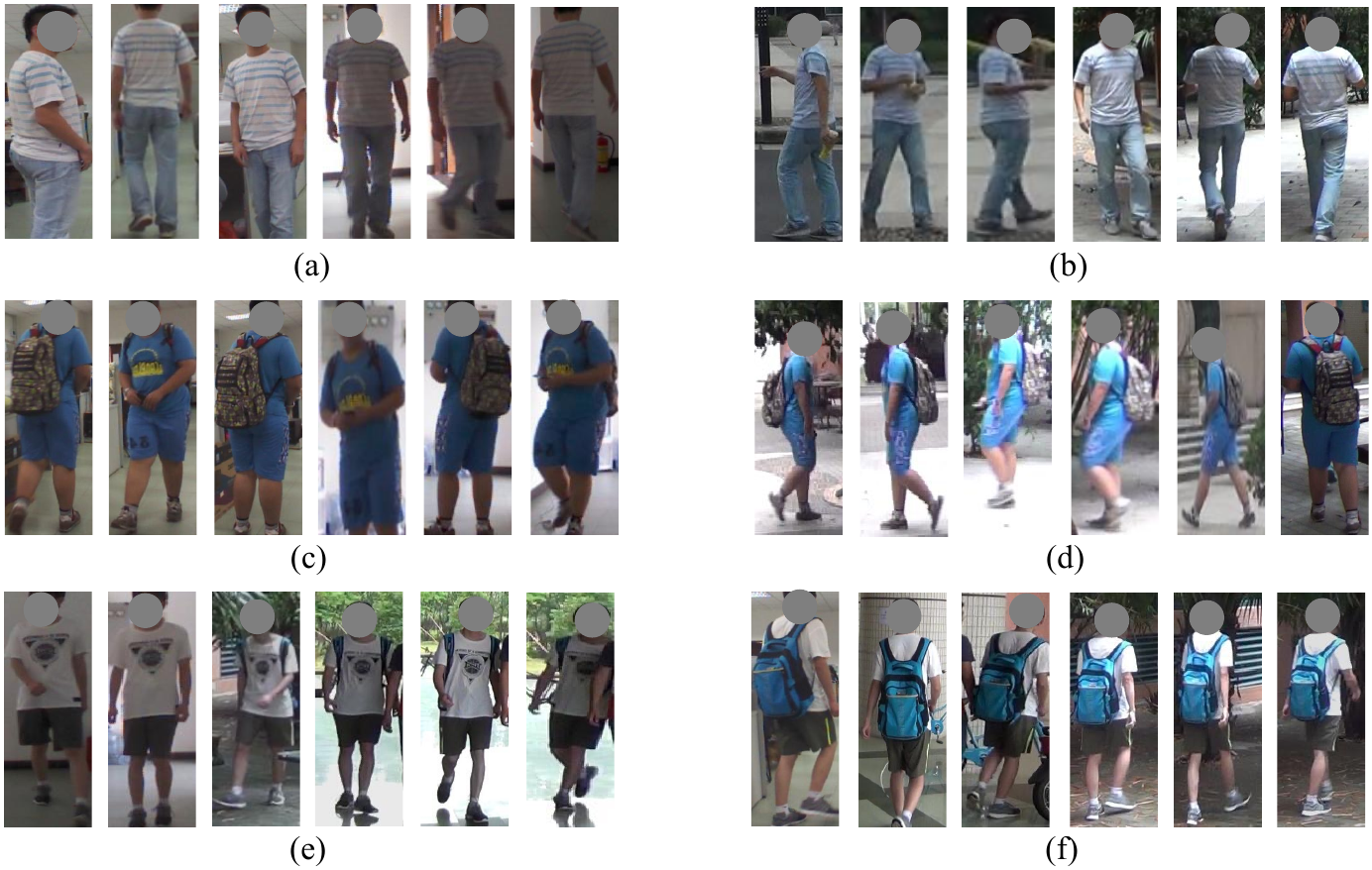}
\caption{Some image examples drawn from the intra-modality and global clustering result. Images in each row belong to the same identity. In each row, the figures on the left and on the right are clustered into different clusters by intra-modality clustering, but correctly associated by global clustering.}
\label{fig_img_example}
\end{figure}

To intuitively analyze the complementary effect between global clustering and intra-modality clustering, we compare their obtained clusters and provide some image examples in Figure \ref{fig_img_example}. One noticeable observation from the figure is that, the global clustering is able to associate the intra-class images that are falsely divided into different intra-modality clusters due to camera or style variation, such as the examples on the left and right side of each row. 

Our hypothesis is that, the proposed distance rectification enhances the association of cross-modality images, which implicitly also facilitates the association of intra-modality images with larger variations. The visual results demonstrates that global clustering effectively complements intra-modality clustering in coping with intra-class variation.

\subsection{A4.3  HyperParameter Analysis}

\subsubsection{Effect of $k_1$ and $k_2$.}
$k_1$ and $k_2$ are two key hyper-parameters in computing the Jaccard distance. To investigate their influence on the model performance, Table \ref{table_k1_k2} provides the results of varying $k_1$ and $k_2$. From the table, we see that the model performance is relatively robust when changing $k_1$ and $k_2$ within reasonable ranges. Specifically, increasing $k_1$ has a bigger impact on All accuracy, while increasing $k_2$ impacts both All and Indoor accuracy, possibly due to the incorporation of too many false positives for distance calibration.

\begin{table}[h]
\centering
\scalebox{0.86}{
\begin{tabular}{cc|cccc}
\toprule
$k_1$ & $k_2$ & R1(\textit{All}) & mAP(\textit{All}) & R1(\textit{Indoor}) & mAP(\textit{Indoor}) \\    
\midrule
30 & 6 & 67.1 & \textbf{63.1} & \textbf{75.0} & \textbf{78.6} \\
40 & 6 & 64.3 & 61.9 & 72.5 & 77.0 \\
20 & 6 & \textbf{67.3} & 59.3 & 69.5 & 73.8 \\
30 & 12 & 64.4 & 62.0 & 71.8 & 76.6 \\
30 & 4 & 67.7 & 62.2 & 74.1 & 77.3 \\
40 & 12 & 62.8 & 60.5 & 72.6 & 77.6 \\
\bottomrule                
\end{tabular}
}
\caption{Analysis on $k_1$ and $k_2$.}
\label{table_k1_k2}
\end{table}

\subsubsection{Effect of \textit{Eps}.}
\begin{figure}[h]
\centering
\includegraphics[width=0.39\textwidth]{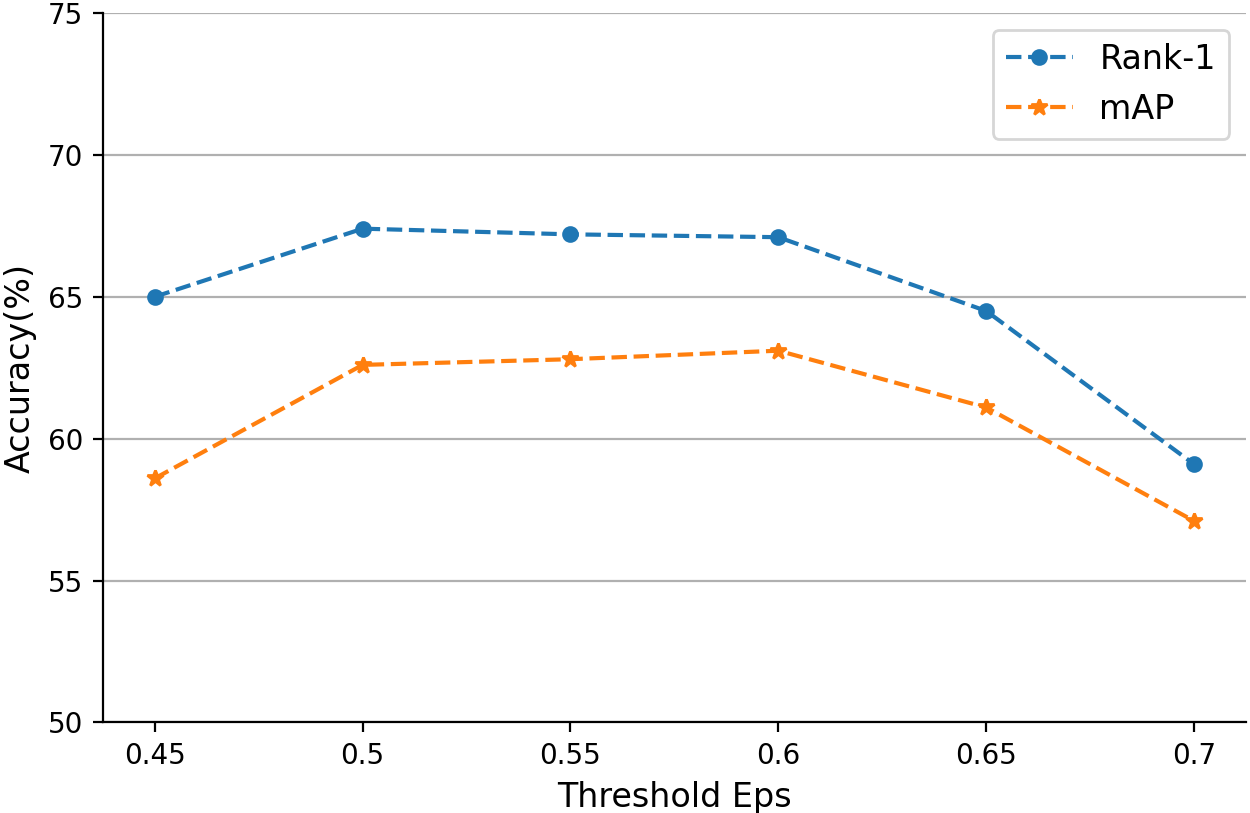}
\caption{Analysis on threshold \textit{Eps}.}
\label{fig_eps}
\end{figure}

The eps value for clustering are set by following existing state-of-the-art USVI-ReID methods including PGM~\cite{PGMAL}, MMM and PCLHD. Figure \ref{fig_eps} presents our model's accuracy on SYSU-MM01 with varying eps values. From the table, we see that the model retains a relatively good performance when eps changes from 0.5 to 0.65, and the best performance is achieved when \textit{Eps} is 0.6.

\subsubsection{Effect of temperature $\tau$.}
From Table \ref{table_tau}, the temperature has a bigger impact on the performance, and 0.05 works best, which is also a common choice adopted by many USVI-ReID methods.

\begin{table}[h]
\centering
\scalebox{0.87}{
\begin{tabular}{c|cccc}
\toprule
$\tau$ & R1(\textit{All}) & mAP(\textit{All}) & R1(\textit{Indoor}) & mAP(\textit{Indoor}) \\    \midrule
0.03 & 64.5 & 59.4 & 72.1 & 76.0 \\
\textbf{0.05} & \textbf{67.1} & \textbf{63.1} & \textbf{75.0} & \textbf{78.6} \\
0.07 & 60.3 & 54.7 & 65.5 & 71.3 \\
\bottomrule                
\end{tabular}
}
\caption{Analysis on temperature $\tau$.}
\label{table_tau}
\end{table}

\subsection{A4.4  Analysis on Subset Clustering}

\subsubsection{The subset sampling ratio.}
\begin{figure}[h]
\centering
\includegraphics[width=0.37\textwidth]{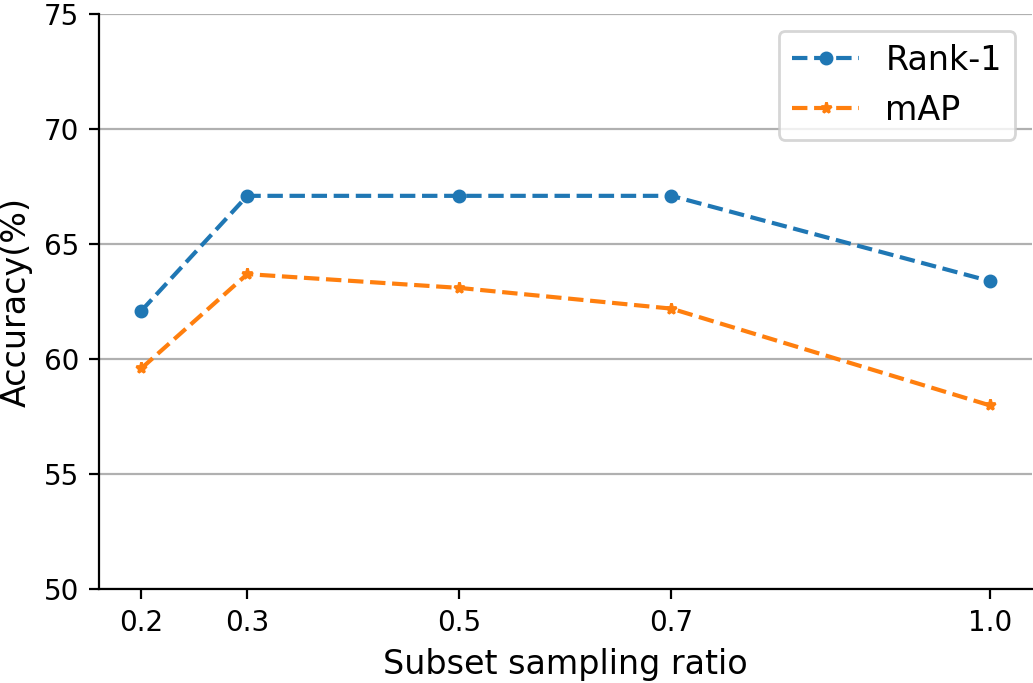}
\caption{Accuracy under different subset sampling ratio.}
\label{fig_sample_ratio}
\end{figure}

Subset sampling ratio controls the number of visible images sampled for intra-modality and global clustering. To investigate how the sampling ratio influences model performance, we vary the ratio from 0.2 to 1, and train our model at each ratio respectively. Figure \ref{fig_sample_ratio} shows the accuracy curve as the sampling ratio changes. From the figure, we see that the model is able to maintain a good performance when the sampling ratio is within the range of [0.3, 0.7], showing that our method is consistent and robust under a wide range of sampling ratios. When setting the ratio to 1, \textit{i.e.}, without subset sampling, the model suffers from an obvious accuracy drop, proving the effectiveness and necessity of the subset sampling strategy.

\begin{figure}[h]
\centering
\includegraphics[width=0.42\textwidth]{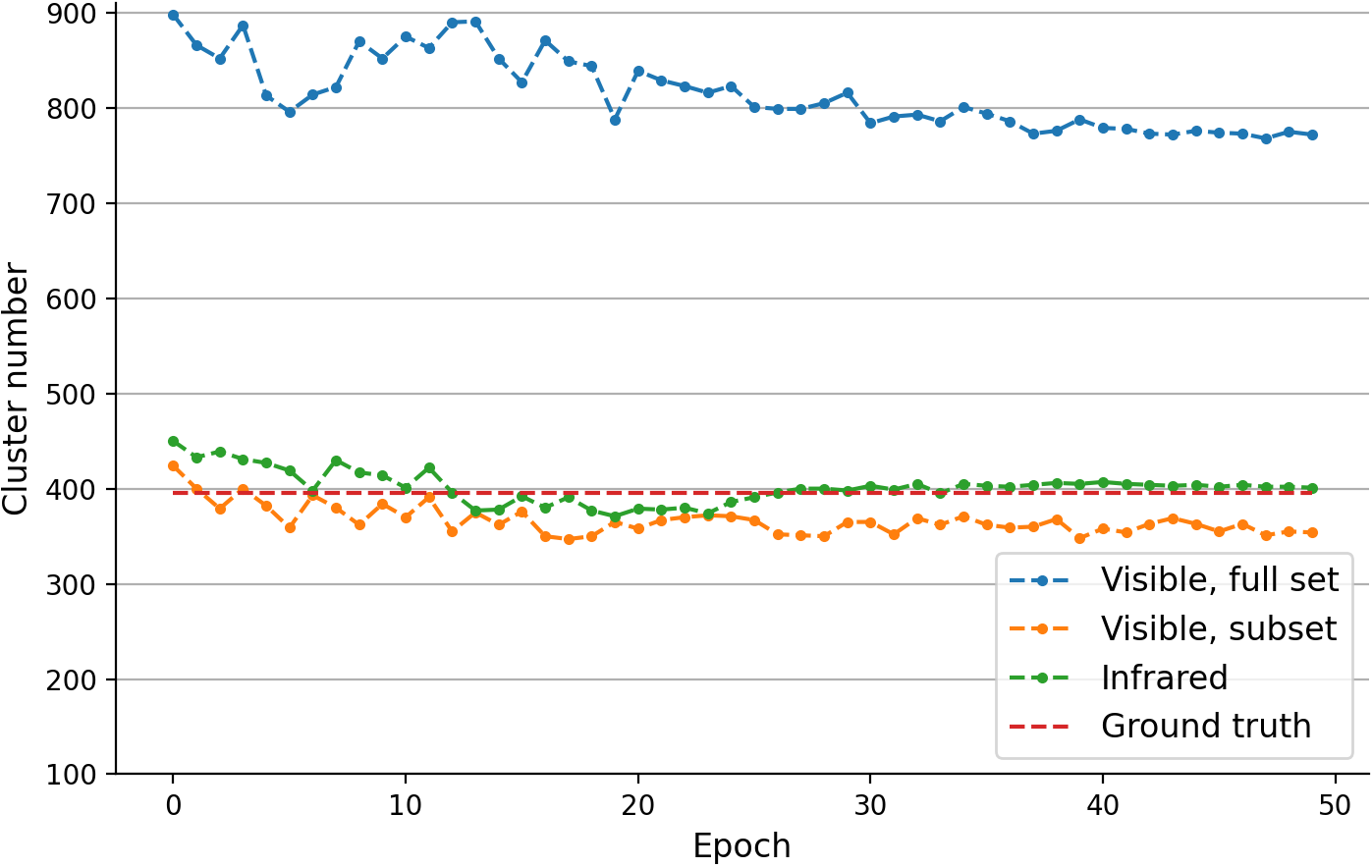}
\caption{Cluster number evolvement.}
\label{fig_cluster_num}
\end{figure}

\subsubsection{Full set \textit{v.s.} subset clustering.}
To investigate the impact of subset clustering, Figure \ref{fig_cluster_num} plots the number of clusters predicted by full set and subset clustering respectively, compared to the ground truth. 

First, we observe that the visible modality demonstrates clear over-clustering, generating much more clusters compared to the ground truth. This is mainly due to the merging-based clustering algorithm (\textit{e.g.} DBSCAN) cannot well handle datasets with too many images per class. And the visible modality of SYSU-MM01 has almost twice number of images compared to the infrared modality. 

Second, we can see that utilizing subset clustering on the visible modality effectively alleviates over-clustering. The cluster number predicted by subset clustering is closer to the ground truth, and shows a \textbf{\textit{relatively stable evolvement}} along the training.

\subsubsection{Effect of Subset Clustering on RegDB.}
The subset clustering strategy is not intended to be a universal strategy, but rather targeted at datasets with a large number of per-class images, which can be very common in real-world applications. Datasets like RegDB with \textit{limited and balanced} number of images across modalities do not suffer from over-clustering, therefore the subset clustering strategy may not have a positive effect in such scenarios. For reference, we provide the results of utilizing subset clustering on RegDB, as summarized in Table \ref{table_regdb}. From the Table, we see that utilizing subset clustering on RegDB yields a slightly lower, yet still competitive accuracy compared to full-set clustering.

\begin{table}[h]
\centering
\scalebox{0.9}{
\begin{tabular}{c|cccc}
\toprule
Setting & R1(\textit{V2I}) & mAP(\textit{V2I}) & R1(\textit{I2V}) & mAP(\textit{I2V}) \\    \midrule
Full set & 94.3 & 89.1 & 93.6 & 88.5 \\
Subset  & 93.6 & 88.6 & 93.0 & 88.2  \\    
\bottomrule                
\end{tabular}
}
\caption{Comparison of full set and subset clustering on RegDB. \textit{V2I}: Visible to Thermal. \textit{I2V}: Thermal to Visible.}
\label{table_regdb}
\end{table}

\end{document}